\documentclass[times,twocolumn,final]{elsarticle}
\usepackage[top=0.85in, bottom=0.95in, left=0.75in, right=0.75in]{geometry}

\makeatletter
\def\ps@pprintTitle{%
 \let\@oddhead\@empty
 \let\@evenhead\@empty
 \let\@oddfoot\@empty
 \let\@evenfoot\@empty
}
\makeatother

\usepackage{setspace}
\usepackage{parskip}
\usepackage{etoolbox}

\usepackage{booktabs}
\usepackage{multicol}
\usepackage{afterpage}
\usepackage{placeins}
\usepackage{ltablex}
\usepackage{caption}
\usepackage{supertabular}
\usepackage[flushleft]{threeparttable}
\usepackage{dblfloatfix}
\usepackage{framed,multirow}

\usepackage{amssymb}
\usepackage{latexsym}

\usepackage{url}
\usepackage{xcolor}  
\usepackage{hyperref}  
\usepackage{amsthm}
\theoremstyle{definition}
\newtheorem{definition}{Definition}[section]
\usepackage{amsmath}
\usepackage{algorithm}
\usepackage{algorithmic}
\usepackage{amssymb}
\usepackage{moreverb}
\usepackage{latexsym}
\usepackage{graphicx}
\usepackage{longtable}
\usepackage{afterpage}

\begin{document}


\begin{frontmatter}

\title{Deep Learning Approaches for Medical Imaging Under Varying Degrees of Label Availability: A Comprehensive Survey}%

\author[label1]{Siteng Ma\corref{cor1}}
\author[label1]{Honghui Du\corref{cor1}}
\author[label1]{Yu An\fnref{equal1,ucd1}}
\author[label2]{Jing Wang\fnref{equal1}}
\author[label3]{Qinqin Wang\fnref{equal1,ucd1}}
\author[label4]{Haochang Wu}
\author[label1]{Aonghus Lawlor}
\author[label1]{Ruihai Dong}
\cortext[cor1]{Corresponding authors: Siteng Ma \textless{}siteng.ma@insight-centre.org\textgreater{}, 
Honghui Du \textless{}honghui.du@insight-centre.org\textgreater{}}
\fntext[equal1]{These authors contributed equally to this work.}
\fntext[ucd1]{This work was conducted while the authors were affiliated with University College Dublin.}

\address[label1]{The Insight Centre for Data Analytics, School of Computer Science, University College Dublin, Dublin, Ireland}
\address[label2]{College of Computer Science, North China Institute of Aerospace Engineering, Langfang, China P.R.}
\address[label3]{Hithink RoyalFlush Information Network Co., Ltd, Hangzhou, China}
\address[label4]{School of Electrical and Electronic Engineering, University College Dublin, Dublin, Ireland}

\begin{abstract}

Deep learning has achieved significant breakthroughs in medical imaging, but these advancements are often dependent on large, well-annotated datasets. However, obtaining such datasets poses a significant challenge, as it requires time-consuming and labor-intensive annotations from medical experts. Consequently, there is growing interest in learning paradigms such as incomplete, inexact, and absent supervision, which are designed to operate under limited, inexact, or missing labels. This survey categorizes and reviews the evolving research in these areas, analyzing around 600 notable contributions since 2018. It covers tasks such as image classification, segmentation, and detection across various medical application areas, including but not limited to brain, chest, and cardiac imaging. We attempt to establish the relationships among existing research studies in related areas. We provide formal definitions of different learning paradigms and offer a comprehensive summary and interpretation of various learning mechanisms and strategies, aiding readers in better understanding the current research landscape and ideas. We also discuss potential future research challenges. 

\end{abstract}

\begin{keyword}
Deep learning \sep Active learning \sep Semi-supervised learning \sep Inexact learning \sep Unsupervised learning \sep Transfer learning \sep Medical image analysis
\end{keyword}

\end{frontmatter}


\section{Introduction}
Medical imaging plays a crucial role in clinical decision-making and represents a dominant modality in healthcare, encompassing various techniques such as ultrasound (US), X-ray, Computed Tomography (CT), and Magnetic Resonance Imaging (MRI). The data extracted from these imaging methods have a wide range of clinical applications, including computer-aided diagnosis, treatment planning, intervention, and therapy \citep{bi2019artificial}. However, interpreting such complex medical images is a skill-intensive task, typically requiring qualified radiologists to undergo extensive training and accumulate substantial experience \citep{royal2016clinical}.

\begin{figure*}
\centering
\includegraphics[width=0.78\textwidth]{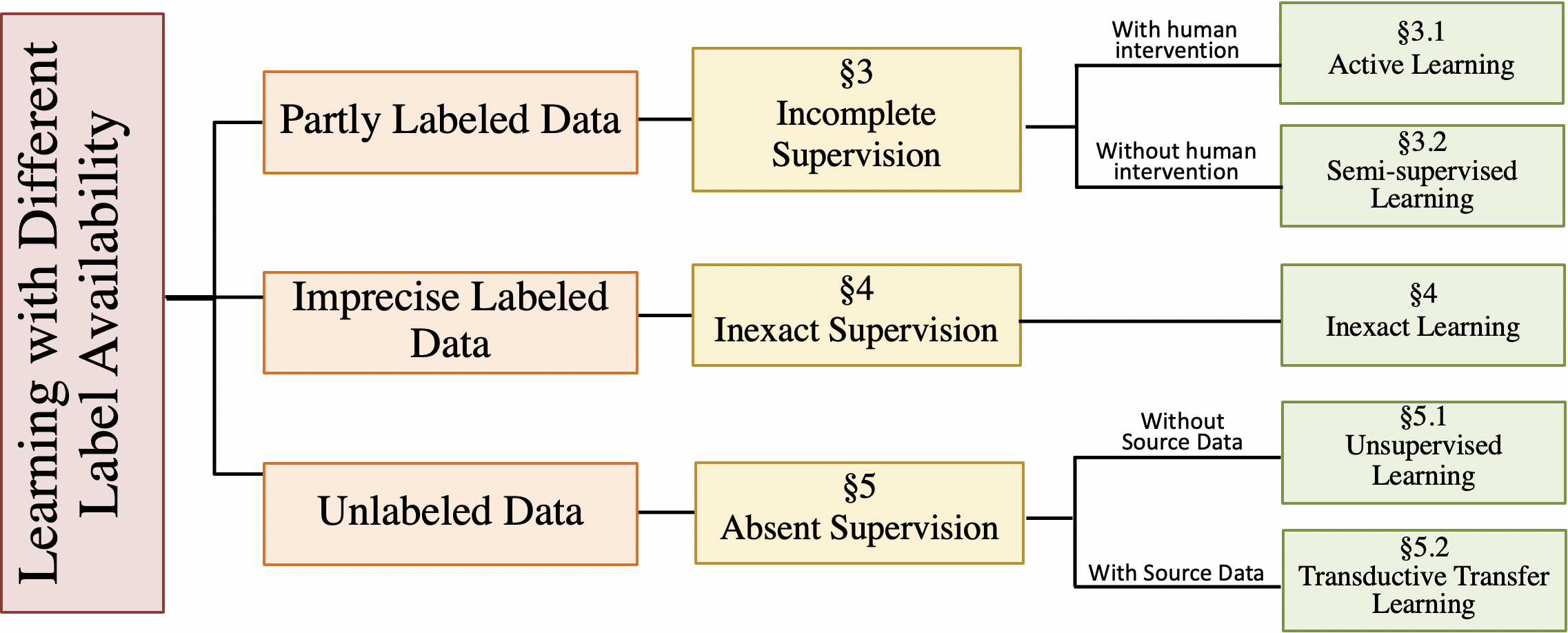}
\caption{Mindmap of Learning Patterns Under Varying Degrees of Label Availability. }
\label{fig:Overall_tree}
\end{figure*}
Deep learning (DL) has become widely utilized in medical image analysis due to its capability to automatically learn complex patterns and features from large volumes of labeled data, leading to more efficient diagnosis and analysis compared to traditional manual methods \citep{luo2021semi}. Nevertheless, the annotation of large datasets remains challenging, as manual slice-by-slice labeling is costly, time-consuming, and heavily reliant on expert experience \citep{top2011active}. 

To address the challenge posed by the severe scarcity of labeled data, one potential solution is to enable DL models to learn not only from the limited available labeled data but also from substantial amounts of easily accessible unlabeled or imprecisely labeled data. Therefore, alternative learning paradigms, such as incomplete, inexact, and absent supervision, have gained growing attention in medical image analysis over the past decade.

While several recent surveys have addressed related topics, their scopes have generally been limited. For instance, \cite{budd2021survey} concentrated exclusively on deep active learning, while \cite{shurrab2022self} focused solely on semi-supervised learning in medical applications. Other surveys have targeted specific medical domains, such as Covid-19 imaging \citep{shorten2021deep} or breast cancer imaging \citep{debelee2020survey}, further restricting their applicability. Two influential surveys closely related to our work include \cite{cheplygina2019not}, which provided a comprehensive review of semi-supervised, multi-instance, and transfer learning but was confined to literature before 2018, and although \cite{tajbakhsh2020embracing} covered a broader range of scenarios, but focused exclusively on segmentation tasks up to 2019. Distinctively, our survey advances beyond these limitations by offering the following detailed and novel contributions:

\begin{enumerate}
\item We comprehensively integrate three major paradigms covering five scenarios, including active and semi-supervised learning (incomplete supervision), inexact supervised learning (inexact supervision), and unsupervised and transductive transfer learning (absent supervision), clearly elucidating their interactions, commonalities, and differences. This integration provides an unprecedented, unified framework for researchers. The overall review structure can be seen in Fig. \ref{fig:Overall_tree}.
\item We examine multiple medical imaging tasks (classification, segmentation, and object detection) across a wide range of imaging modalities (MRI, CT, X-ray, US, and more). This comprehensive analysis enables a deeper understanding of how tasks, datasets, anatomical targets, and modalities align with different learning paradigms, thereby facilitating methodological transfer and paradigm selection across diverse clinical imaging scenarios.
\item Our work provides a comprehensive analysis of recent methodological developments, highlights emerging challenges and trends, and outlines key open questions, offering researchers valuable, forward-looking guidance.

\end{enumerate}

\begin{table}[h]
\centering
\caption{List of Abbreviations Used in This Paper}
\label{tab:abbreviations}
\resizebox{0.75\linewidth}{!}{  
\begin{tabular}{ll}
\hline
\textbf{Abbreviation} & \textbf{Full Term} \\
\hline
ISL & Incomplete Supervised Learning \\
AL & Active Learning \\
Semi-SL & Semi-supervised Learning \\
ISL & Inexact supervised Learning \\
UL & Unsupervised Learning \\
Self-SL & Self-supervised Learning \\
CL & Contrastive Learning \\
TTL & Transductive Transfer Learning \\
PL & Pseudo Labeling \\
Align & Alignment \\
Cls & Classification \\
Det & Detection \\
Reg & Registration \\
Seg & Segmentation \\
Abd & Abdomen \\
Derm & Dermatology \\
Endo & Endoscopy \\
GI & Gastrointestinal \\
Pancre & Pancreas \\
Histo & Histopathology \\
OCT & Optical Coherence Tomography \\
\hline
\end{tabular}}
\end{table}

In this survey, we reviewed 610 peer-reviewed papers published between 2018 and 2024, sourced from Web of Science (WOS) \footnote{https://clarivate.com} and Scopus \footnote{https://www.elsevier.com/products/scopus} databases, based on clearly defined search criteria available at https://github.com/HelenMa9998/Awesome-Varying-Degrees-of-Label-for-Medical-Image-analysis. These papers are published in top medical and computer vision journals and conferences, including but not limited to Medical Image Analysis (MIA), IEEE Transactions on Medical Imaging (TMI), the International Conference on Medical Image Computing and Computer-Assisted Intervention (MICCAI), the International Conference on Bioinformatics and Biomedicine (BIBM), the IEEE/CVF Conference on Computer Vision and Pattern Recognition (CVPR), the International Conference on Computer Vision (ICCV), and the European Conference on Computer Vision (ECCV).  Due to space constraints, we provide an in-depth analysis of 172 key papers within this manuscript, while the complete list is detailed in the supplementary material.

\begin{figure*}
\centering
\includegraphics[width=0.9\textwidth]{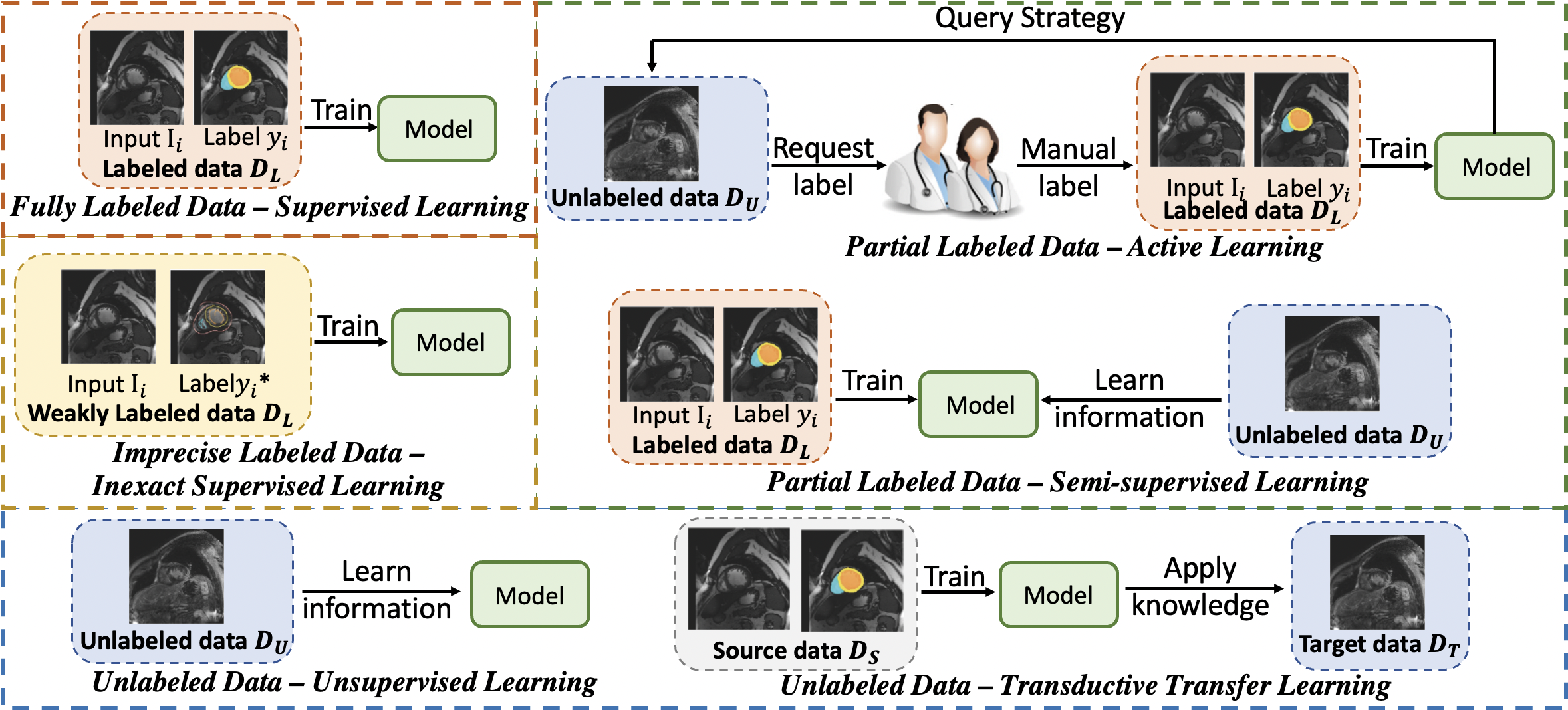}
\caption{\label{fig:all_method_illustration}Here we use the Cardiac MRI segmentation task as an example to show different learning scenarios. In supervised learning, fully labeled data is given. In active learning (Section \ref{Active Learning (AL)}), a small amount of labeled data is combined with a large amount of unlabeled data and human annotation is used as additional assistance. Semi-supervised learning (Section \ref{Semi-supervised Learning (Semi-SL)}) is a similar scenario. In inexact supervised learning (Section \ref{sec: Inexact Supervision}), only scribble labels are available. Unsupervised learning (Section \ref{Unsupervised Learning}) only involves unlabeled data. Transductive transfer learning (Section \ref{Transductive Transfer Learning}) focuses on the labeled data from a different domain, and we use Cardiac MRI from another domain as an example here.}
\end{figure*}

By reviewing and analyzing those relevant papers, this survey provides a comprehensive roadmap for researchers to navigate the rapidly evolving landscape of medical imaging under limited supervision, offering practical insights and identifying opportunities for future innovation. We summarize all abbreviations used throughout the paper in Table~\ref{tab:abbreviations} for clarity. 

This paper is organized as follows: Section \ref{sec: Medical Image Analysis Tasks and Learning Paradigms} defines medical image analysis tasks and machine learning paradigms covered in this survey. Sections \ref{sec: Incomplete Supervision}, \ref{sec: Inexact Supervision}, and \ref{sec: Absent Supervision} introduces learning approaches with limited labeled data, imprecise labels, and unlabeled data, respectively. Section \ref{sec: Discussion} analyses and discusses the trends, challenges, and future directions in this research area. Section \ref{sec: Conclusion} summarizes the key findings of this survey.

\section{Medical Image Analysis Tasks and Learning Paradigms}
\label{sec: Medical Image Analysis Tasks and Learning Paradigms}

Let $\mathbf{I} \in \mathbb{R} ^{h \times w \times d}$ represents a volumetric medical image, where $h\times w$ is the spatial size and $d$ is the number of slices ($d=1$ for 2D images). This survey includes three key tasks in medical image analysis: classification, segmentation, and object detection. 

Classification categorizes whole images into predefined classes based on content \citep{litjens2017survey}, such as distinguishing between benign and malignant tumors in mammograms, or identifying different types of lung diseases in chest X-rays \citep{li2019benign}. It can be defined as:
\begin{definition}[Classification]
    Classification refers to a function $f$ that maps an input $\mathbf{I}$ to a class label $y$ from a predefined set $\mathcal{Y}$, where $y$  represents the image’s overall category, such as the type of disease depicted; $\mathcal{Y}$ contains all possible labels that $\mathbf{I}$ can be categorized into.
\end{definition}

Segmentation divides an image into meaningful regions, such as tissues, organs, or pathologies, to aid diagnosis and treatment planning (e.g., delineating tumors in MRI scans) \citep{liu2021review, ghose2012survey}. Formally, it is defined as:
\begin{definition}[Segmentation]
    A function $f$ maps $\mathbf{I}$ to a segmentation mask $\mathbf{M}$, where $\mathbf{M}$ is a matrix of the same size as $\mathbf{I}$. Each element $m$ in $\mathbf{M}$ assigns a label to the corresponding pixel in $\mathbf{I}$, classifying it into predefined categories like anatomical structures or pathologies.
\end{definition}

Object detection identifies and localizes entities like tumors, fractures, or organs in medical images (e.g., X-rays, MRI, CT) \citep{litjens2017survey}. It can be defined as:
\begin{definition}[Object Detection]
    Object detection is a function $f$ that maps an $\mathbf{I}$ to a set of bounding boxes $\mathcal{B}$ and associated class labels $\mathcal{Y}$, where each box $b_j \in \mathcal{B}$ corresponds to a region of interest, and each label $y_j \in \mathcal{Y}$ specifies the category of the detected feature. 
\end{definition}


Fully labeled datasets required for training $f$ are often unavailable in practice \citep{liu2019weakly}. This study primarily investigates three learning paradigms, namely, incomplete, inexact, and absent supervision, under varying levels of label availability (see Fig. \ref{fig:all_method_illustration}). For clarity, the definitions are illustrated in the context of a classification task. Let $\mathcal{D}_L = \{(\mathbf{I}_i, y_i)\}_{i=1}^{|\mathcal{D}_L|}$ denote a labeled dataset, and $\mathcal{D}_U = \{\mathbf{I}_i\}_{i=|\mathcal{D}_L|+1}^{|\mathcal{D}_U|}$ an unlabeled dataset, where $y_i \in \mathcal{Y}$ is the label corresponding to input $\mathbf{I}_i$. Traditional supervised learning utilizes $\mathcal{D}_L$ to train a model $f(\cdot)$ such that $f(\mathbf{I}_i) = y_i$.

Incomplete supervision occurs when only a small portion of the data is labeled \citep{zhou2018brief}, such as annotating only some scans in an MRI set \citep{liu2019weakly}. It is defined as follows:
\begin{definition}[Incomplete Supervision] 
    Incomplete supervision involves training on datasets with both labeled and unlabeled instances. Formally, it learns $f(\cdot)$ from $\mathcal{D} = \mathcal{D}_L \cup \mathcal{D}_U$, where $|\mathcal{D}_L| \leq |\mathcal{D}_U|$.
\end{definition}

Inexact supervision uses approximate or imprecise labels \citep{zhou2019weakly}, such as rough tumor boundaries in MRI or broadly classified lesions in X-rays. Formally,
\begin{definition}[Inexact Supervision]
\label{def: Inexact Supervision}
In inexact supervision, $\mathcal{D}_L^* = {(\mathbf{I}_i, y_i^*)}_{i=1}^{|\mathcal{D}_L^*|}$, where the coarse-grained label $y^*_i \approx y_i$, $y^*_i \in \mathcal{Y}^*$ indicates approximate representation of the true label $y_i$. The goal is to learn $f(\cdot)$ from $\mathcal{D}_L^*$.
\end{definition}

Absent supervision involves learning from unlabeled data, such as detecting anomalies in medical scans without prior annotations, by either leveraging inherent image structures \citep{zhou2019weakly} or utilizing labeled source data to aid in understanding the unlabeled target data \citep{pan2009survey}. It is defined as follows:
\begin{definition}[Absent Supervision]
\label{Absent Supervision}
Absent supervision either trains on a target dataset containing only input instances with no labels (e.g., $f(\cdot$) learns from $\mathcal{D}_U$) or leverage labeled source data (e.g., $f(\cdot$) uses $\mathcal{D}_L^S$ to improve learning on $\mathcal{D}^T_U$).
\end{definition}

\section{Incomplete Supervision}
\label{sec: Incomplete Supervision}
Incomplete supervision arises when a subset of data is precisely labeled by experts, while others remain unlabeled due to resource constraints. The challenge lies in effectively utilizing both the limited labeled data and the large amount of unlabeled data to enhance learning. There are two primary techniques for training models under incomplete supervision: active learning (AL) and semi-supervised learning (semi-SL), the former involving human intervention and the latter functioning independently \citep{zhou2018brief}.

\subsection{Active Learning (AL)}
\label{Active Learning (AL)}
Unlike traditional supervised learning, AL reduces labeling costs by selectively querying an \textit{oracle} (e.g., a medical expert) to label the most informative data points, achieving comparable or better performance with fewer labeled examples \citep{settles2009active,ren2021survey}. For instance, in detecting early signs of Alzheimer’s from MRI scans, a DL model is first trained on a small labeled dataset. AL then identifies the most uncertain cases in a larger unlabeled dataset, which are labeled by an expert and used to update the model. This iterative process maximizes learning efficiency with minimal labeled data.

\begin{figure}
\centering
\includegraphics[scale=0.35]{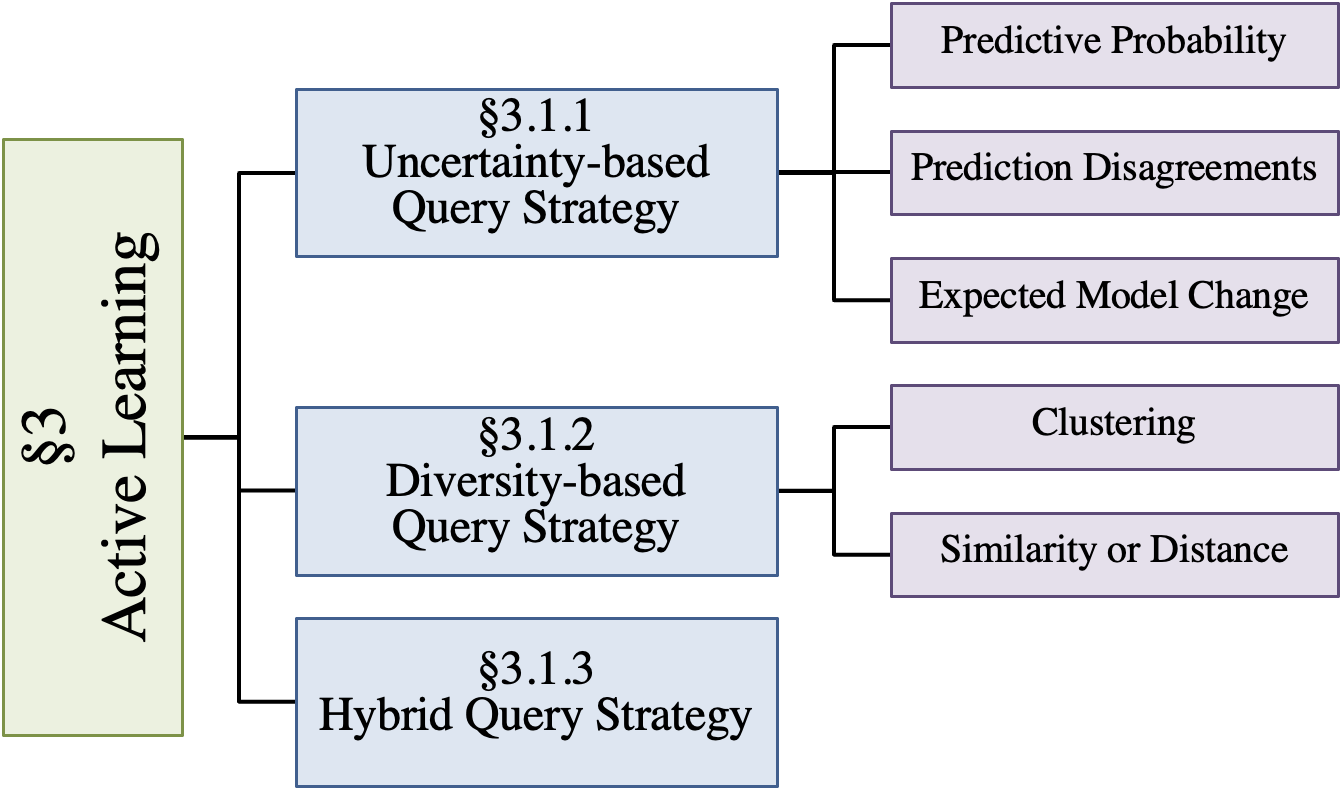}
\caption{\label{fig:Active_tree}The taxonomy of Active Learning. }
\end{figure}

Different AL scenarios (e.g., membership query synthesis, stream-based sampling and pool-based AL) offer varying advantages depending on data and task requirements \citep{settles2009active}. In medical image analysis, most AL research is pool-based. In this survey, AL specifically refers to pool-based AL combined with DL unless stated otherwise. Given definitions in Section \ref{sec: Medical Image Analysis Tasks and Learning Paradigms} and existing AL approaches \citep{ren2021survey}, AL in medical image analysis adopts the following definition:
\begin{definition}[Active Learning]
     An \textit{oracle} $O$ is available to label queried instances from $\mathcal{D}_U$. At each iteration, the AL algorithm selects a subset $\mathcal{D}_Q \subseteq \mathcal{D}_U$ based on a query strategy, sends it to $O$ for labeling, and updates $\mathcal{D}_L = \mathcal{D}_L \cup \mathcal{D}_Q$ and $\mathcal{D}_U = \mathcal{D}_U - \mathcal{D}_Q$. The predictive model $f(\cdot)$ is then trained on $\mathcal{D}_L$. This process repeats until a predetermined stopping criterion is met. The goal is to construct a $f(\cdot)$ that minimizes a given loss function over $\mathcal{D}$ with fewer label queries to $O$.
\end{definition}
There are various query strategies in AL to pick the most informative data points (e.g., uncertainty-, diversity-, and hybrid-based) \citep{budd2021survey, ren2021survey}. Fig. \ref{fig:Active_tree} illustrates the main categories.

\subsubsection{Uncertainty-based Query Strategy}
Uncertainty-based query strategies involve querying the labels of data points for which the model is most uncertain. \textit{Prediction probability} $P(f(\textbf{I}_i))$ derived from the deep neural network's output via the softmax function is often used to directly represent the confidence on the unlabeled instance $\textbf{I}_i \in \mathcal{D}_U$ \citep{hemelings2020accurate}. However, this focuses only on the most likely outcome, ignoring the distribution of probabilities across other possible labels \citep{budd2021survey}. To address this, entropy \citep{shannon2001mathematical} is employed to measure uncertainty across all possible outcomes, reflecting the information needed to represent a distribution \citep{settles2009active}. Entropy is defined as:
\begin{equation}
    H(\mathbf{I}_i) = -\sum_{y \in \mathcal{Y}} P(y|f(\mathbf{I}_i))\log(P(y|f(\mathbf{I}_i)))
\end{equation}
where $P(y|f(\mathbf{I}_i))$ is the probability that the model $f$ predicts label $y$ for sample $\mathbf{I}_i$. \cite{qi2018label,li2021pathal} apply entropy to identify high-uncertainty samples for labeling to solve histopathology tissue classification. However, DL models can sometimes be overly confident in their predictions, leading to an unreliable measure of confidence \citep{ren2021survey}.

\textit{Prediction disagreements} across multiple runs or models is another way to measure uncertainty to avoid over-confidence. Bayesian Active Learning by Disagreement (BALD) \citep{houlsby2011bayesian} treats DL model parameters as probabilistic. For a given input, the model could produce varying predictions across multiple runs. It selects instances with high prediction disagreement to reflect greater uncertainty. In practice, Monte Carlo dropout (MC-dropout) is a commonly used, low-cost method that approximates BALD by applying dropout during inference to generate multiple stochastic forward passes. It has been proven effective in medical imaging tasks such as muscle segmentation \citep{hiasa2019automated} and blood cell detection \citep{sadafi2019multiclass}. \cite{zhao2020deeply,zhao2021dsal} and \cite{zhang2021quality} proposed methods to measure uncertainty by assessing the consistency between segmentation results from the hidden layers and the final layer of a deeply supervised U-Net using Dice’s Coefficient. These methods have shown effective for finger bone segmentation tasks in MRI imaging.


\textit{Expected model change} quantifies uncertainty by estimating the impact an instance would have on the current model (e.g., training loss and gradient) if it were labeled and incorporated into the training set. \cite{dai2020suggestive,dai2022suggestive} proposed a gradient-guided sampling that projects gradients into a variational autoencoder (VAE)'s latent space, identifying instances with higher gradient magnitudes as more uncertain for brain MRI segmentation. 

\subsubsection{Diversity-based Query Strategy}
Although uncertainty helps in selecting samples for labeling, high uncertainty does not always guarantee performance improvement \citep{zhao2020deeply}. Such strategies can bias medical imaging by prioritizing high-uncertainty areas, potentially overlooking critical or underrepresented cases \citep{wang2024comprehensive}. Unlike uncertainty-based strategies, diversity-based query strategies select instances that are both representative and diverse, improving the DL model’s performance across the problem space \citep{budd2021survey}. 

\textit{Clustering} is a common diversity-based strategy. \cite{lin2020two} proposed a hierarchical clustering method for active selection. The method begins by predicting semantic masks for unlabeled images and training two VAE feature extractors separately using segmentation masks and unlabeled images. Initially, mask features are clustered based on appearance similarity. Within each mask cluster, further clustering is performed at the image level. Finally, representative instances from distinct clusters are selected to ensure diversity. 

\textit{Distance or Similarity}-based strategies query samples that are far apart from each other, reducing redundancy and ensuring a more uniform distribution of selected samples. \cite{smailagic2018medal} queries unlabeled instances that maximize the average distance to all labeled samples. \cite{li2023hal} uses cosine distance to select the farthest unlabeled instances from the labeled dataset. Experiments on breast US, liver CT, and chest X-ray segmentation showed high effectiveness. To address significant class imbalance in medical imaging classification, Similarity-based Active Learning (SAL) \citep{zhang2018similarity} learns a similarity score to select diverse samples (distinct from labeled data) and minority-class instances (closely related to labeled rare-class samples) for labeling.

\subsubsection{Hybrid Query Strategy}
Uncertainty-only strategies may redundantly sample similar uncertain instances, while diversity-only strategies might miss highly uncertain samples essential for improving model performance. A hybrid query strategy integrates multiple criteria (e.g., uncertainty and diversity) to overcome the limitations of single-criterion strategies, ensuring they are both informative and diverse \citep{wang2024comprehensive}. 


MedAL \citep{smailagic2018medal} selects unlabeled instances with the highest entropy and greatest feature-space distance from labeled samples for labeling, ensuring both uncertainty and diversity for the data distribution. The Hybrid Active Learning framework using Interactive Annotation (HAL-IA) \citep{li2023hal} quantifies an instance’s informativeness using pixel entropy (PE), region consistency (RC), and image diversity (ID). PE measures uncertainty by averaging pixel entropy, while RC assumes similar grayscale and texture within regions, using SLIC clustering \citep{achanta2012slic}. RC is then computed as the average difference between pixel probabilities and the majority class probability. ID is the minimum cosine distance between an unlabeled and all labeled data. HAL-IA selects samples with the highest PE, RC, and ID. 

Although these methods have demonstrated strong performance, many rely on specifically defined metrics for both uncertainty and diversity, and require careful hyperparameter tuning (e.g., weighting strategies or batch sizes), making them sensitive to dataset characteristics and computationally expensive. Recent approaches \citep{gaillochet2023active,ma2024breaking,ma2024adaptive} aim to achieve hybrid objectives without explicitly defining metrics. For instance, to improve computational efficiency in segmentation by avoiding per-sample uncertainty checks, \cite{gaillochet2023active} randomly divides instances into batches to ensure diversity within each batch, then selects the batch with the highest mean uncertainty for labeling to prioritize valuable instances. \cite{ma2024breaking} further argues that conventional uncertainty-based strategies often suffer from redundancy and inaccuracy, and sometimes underperform random selection in medical segmentation. Therefore, a selective uncertainty-based approach employs thresholding to emphasize uncertainty in target regions and near the decision boundary, combining diversity in an efficient way and yielding significant improvements and substantially surpassing random selection across various segmentation tasks. 

DL models typically begin by learning simple patterns and gradually progress to harder data \citep{arpit2017closer}. However, most deep AL methods use fixed strategies and fail to adapt to these phases. The Adaptive Curriculum query strategy for Active Learning (ACAL) \citep{ma2024adaptive} is the first method to adjust its query strategy based on the DL model’s learning stages. It starts with diversity-based sampling to select instances that cover various difficulty levels and general characteristics. Once the selected data distribution aligns with the overall dataset, ACAL transitions to uncertainty-based sampling, focusing on hard-to-classify samples to progressively train the model from easier to harder data.

\begin{table*}[]
\centering
\caption{Overview of Active Learning Strategies in Medical Imaging Analysis.}
\label{tab:al}
\scalebox{0.88}{
\begin{tabular}{llllll}
\hline
Reference & Category & Sub-category & Body & Modality & Task \\ \hline
\cite{hiasa2019automated} & Uncertainty & Entropy & Muscle & CT & Seg \\
\cite{sadafi2019multiclass} & Uncertainty & Entropy & Cell & Histo & Det \\
\cite{zhao2020deeply} & Uncertainty & Prediction Disagreement & Bone & X-Ray & Seg \\
\cite{zhang2021quality} & Uncertainty & Prediction Disagreement & Brain & MRI & Seg \\
\cite{dai2022suggestive} & Uncertainty & Expected Model Change & Brain & MRI & Seg \\
\cite{dai2020suggestive} & Uncertainty & Expected Model Change & Brain & MRI & Seg \\
\cite{lin2020two} & Diversity & Cluster & Cell & Histo & Seg \\
\cite{smailagic2018medal} & Hybrid & Entropy + Similarity & Eye/Cells/Skin & OCT/Histo/Derm & Cls \\
\cite{gaillochet2023active} & Hybrid & Batch-level Entropy & Prostate & MRI & Seg \\
\cite{ma2024breaking} & Hybrid & / & Brain; Spleen & MRI; CT & Seg \\
\cite{ma2024adaptive} & Hybrid & / & Brain; Cell & MRI; Histo & Cls \\
\cite{qi2018label} & Uncertainty (Semi-SL) & Entropy & Tissue & Histo & Cls \\
\cite{li2021pathal} & Uncertainty (Semi-SL) & Entropy & Tissue & Histo & Cls \\
\cite{ma2023adaptive} & Uncertainty (Semi-SL) & Adversarial Attack & Eye; Cell & Fundus; Histo & Cls \\
\cite{zhao2021dsal} & Uncertainty (Semi-SL + ISL) & Prediction Disagreement & Bone/Skin & X-Ray/Derm & Seg \\
\cite{tang2023pldc} & Uncertainty (Semi-SL) & Prediction Disagreement & Vessel & US & Seg \\
\cite{nath2022warm} & Uncertainty (Semi-SL) & Prediction Disagreement & Liver; Vessel & CT & Seg \\
\cite{hochberg2022self} & Diversity (UL) & Similarity & Chest/Liver & CT & Cls \\
\cite{zhang2018similarity} & Diversity (Semi-SL) & Similarity & GI & Endoscopy & Cls \\
\cite{ma2024model} & Hybrid (Semi-SL) & Gradient + Prediction Disagreement & Brain & MRI & Seg \\
\cite{li2023hal} & Hybrid (ISL) & Entropy + Similarity & Breast/Chest/Liver & US/X-Ray/CT & Seg \\ \hline
\end{tabular}}
\end{table*}

\subsection{Enhancing AL with other techniques}
\label{ALIntegration with other techniques}
AL methods successfully release the labeling burden, but can still suffer from an insufficient exploration of large amounts of unlabeled data. To address these limitations, researchers have begun integrating AL with additional paradigms, such as Semi-Supervised Learning (Semi-SL), Imprecise Supervision (Inexact supervised learning, ISL), and Unsupervised learning, especially Self-Supervised Learning (Self-SL), which will be introduced in the following sections, to enrich training data and further reduce human annotation costs.

\textit{Semi-SL} reduces annotation needs by generating pseudo labels and incorporating them into training, where uncertain data is prioritized for expert annotation, potentially accelerating training but still requiring robust pseudo-label validation. For example, \cite{qi2018label, li2021pathal,zhao2021dsal,tang2023pldc} select high-uncertainty samples for labeling while assigning pseudo labels to low-uncertainty ones. \cite{ma2023adaptive} employs adversarial attacks to identify samples near the decision boundary, assigning pseudo labels to those further away. Class Weights Random Forest \citep{zhang2018similarity} assigns pseudo labels to unlabeled samples resembling labeled data, reducing expert annotation workload.

Semi-SL can further help the AL selection process. Given that carotid US images are highly dependent on user-end protocols and vary significantly in quality, real-world labels on US images typically exhibit high variability. \cite{tang2023pldc} introduces a Pseudo-Label divergence-based Active Learning (PLD-AL) method. PLD-AL operates with two networks: a student network fully trained on the current labeled pool and a teacher network weighted against its prior self and the student. The Kullback-Leibler (KL) divergence between the pseudo-label predictions of the teacher across consecutive AL iterations is used to select samples. Meanwhile, \citep{nath2022warm} addresses the challenge of starting without initial annotated data by proposing a proxy task to generate pseudo labels and using uncertainty metrics to select initial samples with high uncertainty for AL. Subsequently, the approach trains the model using both samples from Semi-SL and AL. Due to the inherent imbalance in tumor and lesion segmentation tasks, \cite{ma2024model} introduces a pseudo-label filter to eliminate potential patches without a target, thereby reducing selection time and allowing the model to focus on more critical regions. \citep{lou2022pixel} used Generative Adversarial Network (GAN) \citep{goodfellow2020generative} to generate more synthetic data with real data for semi-SL to further refine the model for AL selection.

Beyond pseudo-labeling, \textit{ISL} can simplify the manual annotation process to further reduce the labeling burden. \cite{li2023hal} proposes an interactive annotation module with suggested superpixels as a form of weak supervision, allowing guidance to be obtained with just a few clicks, instead of pixel-level labeling.

Meanwhile, \textit{Self-SL} improves model performance by leveraging unlabeled data during pre-training to learn generalizable feature representations. \cite{hochberg2022self} demonstrate this approach by utilizing StyleGAN to learn features through a reconstruction-based pretext task. Unlike conventional AL which iteratively selects and retrains samples, the method employs the Farthest Point Sampling algorithm to select the most distant samples in the latent space, maximizing coverage of data characteristics for one time. This approach reduces the need for retraining and has been validated on medical imaging tasks for Covid-19 and liver tumor classification.  

These studies reflect a growing trend toward hybrid pipelines that combine small-scale, high-quality annotations (AL), with large-scale unlabeled or inexact labeled data. Nevertheless, each hybrid approach needs to carefully balance multiple sources of supervision, which may introduce challenges such as pseudo-label noise, inexact-label imprecision, or increased computational complexity. Addressing these trade-offs remains an important direction for future research.

\paragraph{Summary of Active learning} Table \ref{tab:al} summarizes relevant studies on AL in medical image analysis. The current trend indicates a clear preference for hybrid query strategies that combine uncertainty- and diversity-based strategies, leveraging their complementary strengths. Notably, diversity-based approaches appear less prevalent in medical imaging, possibly due to the subtle inter-class variations typically found in modalities such as CT and MRI, compared to natural images. 

AL offers clear advantages in the medical imaging domain, particularly in reducing annotation costs and making efficient use of limited expert resources. However, it also faces persistent challenges, including performance inconsistencies across tasks, sensitivity to the quality of the initial labeled set, and dependence on well-calibrated sampling strategies \citep{ren2021survey}. Further studies are needed to address these issues.

Meanwhile, from Table \ref{tab:al}, approximately half of the works integrate AL with other techniques (e.g., semi-supervised or inexact supervised methods) to overcome its limited capacity to leverage large-scale unlabeled datasets, reflecting a steadily growing trend in recent years. In practice, these integrated approaches have shown stronger generalization and robustness across various medical domains.

\subsection{Semi-supervised Learning (Semi-SL)}
\label{Semi-supervised Learning (Semi-SL)}
Semi-SL, as another incomplete supervised method, trains models using both labeled and unlabeled data to enhance performance. Unlike AL, it uses unlabeled data without requiring human expert intervention \citep{cheplygina2019not}. Based on Section \ref{sec: Medical Image Analysis Tasks and Learning Paradigms} and \cite{van2020survey}, Semi-SL is defined as follows:
\begin{definition}[Semi-supervised Learning]
     Given a labeled dataset $\mathcal{D}_L$ and an unlabeled dataset $\mathcal{D}_U$, Semi-SL aims to learn a model $f(\cdot)$ from $\mathcal{D} = \mathcal{D}_L \cup \mathcal{D}_U$ by leveraging both labeled and unlabeled data to improve generalization.
\end{definition}

As Semi-SL uses unlabeled data for learning, it often relies on additional assumptions that the data structure may provide insights for prediction \citep{zhu2009introduction}. The three common assumptions are: \textit{smoothness assumption}, i.e. close instances in input space share the same label; \textit{clustering assumption}, i.e. instances in the same cluster belong to the same class; and \textit{manifold assumption}, i.e. high-dimensional data can be projected onto a lower-dimensional manifold, where points on the same manifold share labels \citep{cheplygina2019not,van2020survey}. 

\begin{figure}
\centering
\includegraphics[width=0.45\textwidth]{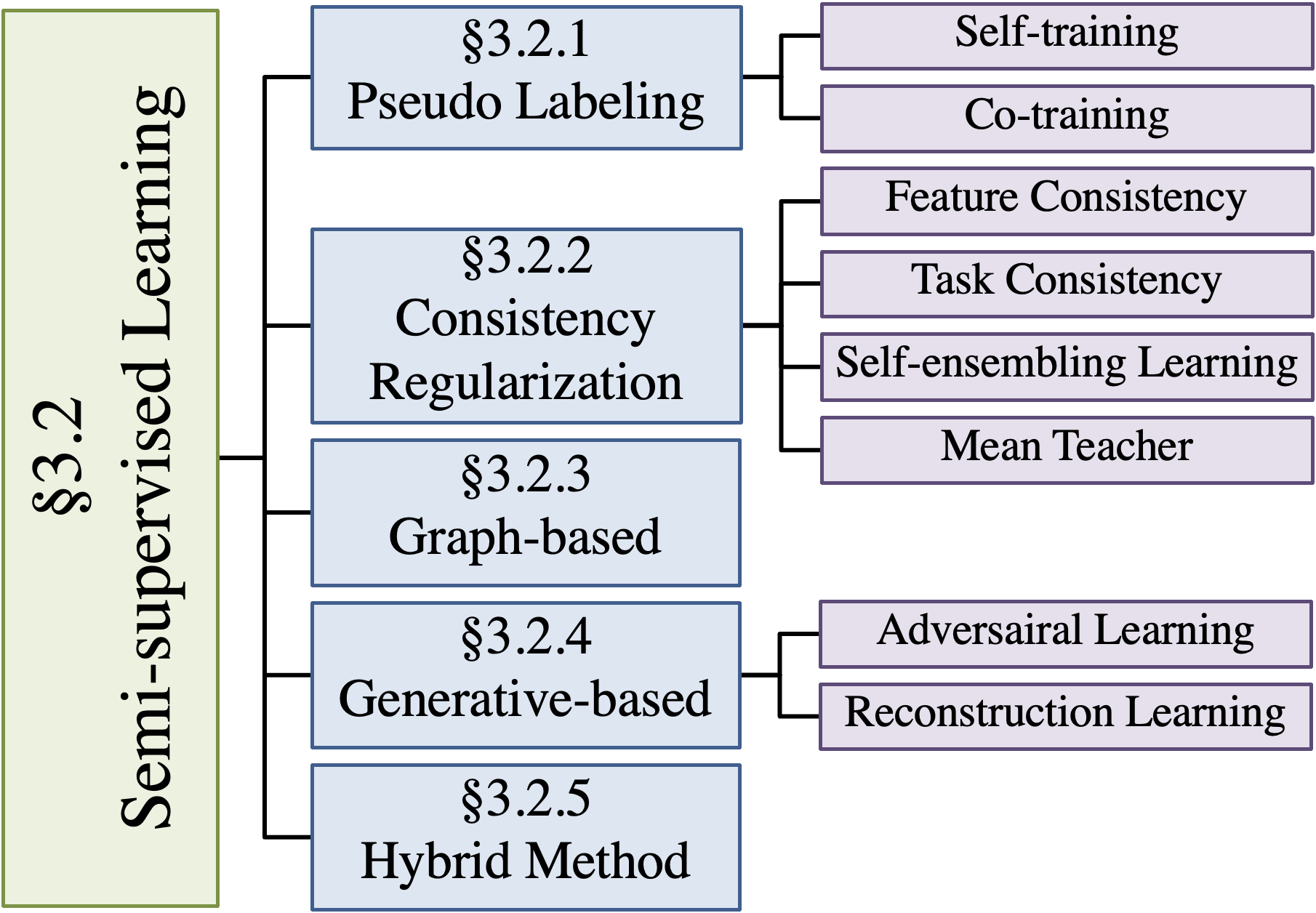}
\caption{\label{fig:SSL_tree}The taxonomy of Semi-supervised Learning. }
\end{figure}

We classify key Semi-SL methods in medical image analysis into five groups: pseudo labeling, consistency regularization, graph-based method, generative method and hybrid method. Fig \ref{fig:SSL_tree} illustrates the primary categories.

\subsubsection{Pseudo Labeling}
\label{Pseudo Labeling}
Pseudo-labeling assigns predicted labels to unlabeled data as ground truth to iteratively train models for improved performance \cite{ito2019semi}. There are two main patterns: self-training, where a single model generates and refines pseudo-labels, and co-training, where multiple models or views generate pseudo-labels for each other to enhance learning.

\textit{Self-training} begins by training a model on a small labeled dataset. The model generates pseudo-labels for unlabeled data, adding high-confidence predictions to the labeled set. It is then retrained iteratively until convergence or a stopping criterion \citep{lee2013pseudo}. The training loss is defined as:
\begin{equation}
L = \frac{1}{|\mathcal{D}_{L}|} \sum_{i=1}^{|\mathcal{D}_{L}|}L({y_{i}},{f^t(\mathbf{I}_i)}) + \frac{\lambda}{|D_{U}|} \sum_{i'=1}^{|\mathcal{D}_{U}|}L(y_{i'}, f^t(\mathbf{I_{i'}}))
\end{equation}
where $t$ is the current timestamp; $y_{i'}$ is the pseudo-label for $\mathbf{I}'$; $\lambda$ balances the supervised and unsupervised loss, and $L(\cdot)$ is the task’s loss function.

The most straightforward approach (i.e., \cite{ito2019semi}) generates pseudo-labels by assigning predictions with confidence exceeding a predefined threshold (e.g., 0.5). However, estimating the optimal value for the predefined threshold is often challenging. \cite{xu2024expectation} automatically and dynamically adjust thresholds under Bayes’ theorem to improve pseudo-label accuracy. Based on the smoothness assumption, \cite{han2022effective} assigns pixel-level pseudo-labels by measuring the distance between pixel representations and class representations, selecting the closest match, and achieving strong performance in liver segmentation. \cite{wu2022minimizing} refine the model’s pixel pseudo-labels by assessing the risk of incorrect predictions, based on the distance between pixel representations and class representations. Mutual Correction Framework (MCF) \citep{wang2023mcf} introduces two different submodels and utilizes the differences between them to correct the model's cognitive bias. It compares the performance of the two submodels on labeled data using the dice loss and selects the better-performing model to generate pseudo-labels, which are then used to guide the training of both submodels.

\textit{Co-training} \citep{huo2021atso, zhu2021semi, wu2023compete, bortsova2019semi} involves training multiple models simultaneously on different views or subsets of the data, in contrast to self-training, which relies on a single model for both pseudo-labeling and learning, potentially leading to more stable training in complex medical imaging datasets. Each model in Co-training generates pseudo-labels for unlabeled data and updates based on pseudo-labels produced by the others, leveraging their agreement on unlabeled data to enhance learning. The process iterates until model performance stops improving \citep{yang2022survey,chen2022semiun}. Yet, these co-training setups often demand carefully chosen model architectures and may increase computational overhead. 

\cite{luo2022semi} is the first to use both CNNs (i.e., Unet \citep{ronneberger2015u}) and Transformers (i.e., Swin-UNet \citep{cao2022swin}) within a co-training framework to capture local features and long-range dependencies, respectively. Uncertainty-guided Collaborative Mean-Teacher (UCMT) \citep{shen2023co} adopts the collaborative mean-teacher paradigm (see Section \ref{Mean Teacher (MT)}) within the co-training framework to support model training. \cite{su2024mutual} improves the reliability of pseudo-labels by comparing the similarity between a feature representation and the average feature representation of the class associated with its pseudo-label. Deep Mutual Distillation (DMD) \citep{xie2023deep} improves segmentation by enabling two networks to mutually learn high-entropy distilled probabilities from each other, allowing them to retain uncertainty in ambiguous regions and reducing the impact of noisy pseudo-labels.

\subsubsection{Consistency Regularisation}
\label{Consistency Regularization}
Pseudo-labeling methods rely on the model's predictions as supervisory signals, which can introduce bias, especially when the model is uncertain or overconfident on unlabeled data. Under the smoothness and clustering assumptions, the output of an input should remain consistent even when subjected to slight perturbations \citep{yang2022survey}. As another method, Consistency regularisation uses labeled data to guide model training while encouraging consistent predictions on unlabeled data under various input perturbations (e.g., geometric transformations, noise injection) or model perturbations (e.g., dropout, weight changes), leveraging both to improve performance \citep{chen2022semiun}.


\textit{Feature consistency} ensures consistent predictions by aligning diverse feature representations to enforce their similarity. Typically, diverse feature representations are generated by feeding images or their augmentations into multiple networks \citep{berenguer2024semi,wu2022mutual,peiris2021duo} or multiple scales (different layers of one network) for a given input \citep{luo2021efficient, luo2022semi1}. 

Recent works \citep{xia20203d,berenguer2024semi,liu2021federated,xu2019deepatlas} underscore a shift from simple data-augmentation or network-based consistency to more sophisticated multi-modal and multi-task alignment, a promising direction that could further enhance the robustness of medical image analysis models. For example, Uncertainty-aware Multi-view Co-Training (UMCT) \citep{xia20203d} promotes diversified features by generating different views through rotation or permutation of 3D data and employing asymmetrical 3D kernels across different sub-networks. \citep{berenguer2024semi} uses two different networks to generate the feature from two augmented views of the same image and minimizes the distance between these features for medical image classification. Additionally, it explored graph-based structures to further align the relations between features from each view by cross-entropy to enforce consistency. Federated Semi-Supervised Learning \citep{liu2021federated} incorporates Semi-SL to train a federated model by applying consistency regularization to the diverse feature representations of each example from unlabeled clients (e.g., hospitals). \cite{chen2022mass} extracts independent features from different modalities (e.g., CT and MRI). It employs consistency regularization in both the semantic and feature spaces to distill cross-modal consistency and ensure feature alignment across these modalities. Diverse feature representations can be generated from different tasks \citep{xu2019deepatlas}, where registration, which aligns two or more images into a common coordinate system, is used to support the segmentation by penalizing the discrepancies between the segmentation of the moving image and that of the target image.


\textit{Task consistency} usually trains a single encoder to perform multiple related tasks simultaneously and enforce consistent predictions across multiple related tasks, helping to generate more accurate pseudo labels for the target task. The shared representations regulate the model and enhance its generalization. \cite{luo2021semi} presents a model that fosters consistency between image-level set functions (a mathematical framework defines and tracks object contours) \citep{ma2020learning}, and segmentation tasks. Using the dual task consistency regularization, this approach incorporates geometric constraints and leverages unlabeled data to enhance the training process. Similarly, \citep{chen2022semi} adopts a dual-branch framework to predict both segmentation and signed distance maps (distance to the nearest boundary of a segmentation object). For unlabeled data, a consistency loss fosters mutual learning between the two tasks.


\textit{Self-ensembling Learning} involves making multiple predictions on unlabeled data, and averaging the prediction results to generate a more robust model prediction in a single model \citep{laine2016temporal}. There are similarities between self-ensembling and feature consistency, as they both combine different variants of the data. However, self-ensembling emphasizes model diversity during training by updating parameters across iterations or epochs and fusing them to enhance robustness in both the model and its predictions. For example, Temporal Ensembling (TE) averages predictions across epochs, and stabilizes learning outputs over time \citep{laine2016temporal}. However, fluctuations in predictions from individual cycles can introduce bias, especially when the model performs poorly. To address this, \cite{cao2020uncertainty} introduced uncertainty weight when calculating the average of cumulative predictions, to assign greater weight to more confident model outputs, and less to uncertain outputs. 

\label{Mean Teacher (MT)}
\textit{Mean Teacher (MT)} framework \citep{tarvainen2017mean} builds on the concept of TE by incorporating Exponential Moving Average (EMA) for model weight updates. Unlike epoch-wise averaging prediction in TE, EMA updates the teacher model's weights incrementally from students after each training iteration, enabling more frequent and consistent alignment with the student model. The MT utilizes two networks with identical architectures but distinct parameters: a student network (\(\theta\)) and a teacher network (\(\theta'\)). The teacher network is updated using EMA of the student network's parameters:
\begin{equation}
\theta'_{t} = \alpha \cdot \theta'_{t-1} + (1 - \alpha) \cdot \theta_t
\end{equation}
where \(\alpha\) is the smoothing coefficient and \(t\) denotes the training iteration. The loss function for training combines two terms:
\begin{align}
L_{MT} &= L_{cls} + L_{con} \notag = \frac{1}{|\mathcal{D}_L|} \sum_{i=1}^{|\mathcal{D}_L|} L(y_{i}, f_{\theta}(\mathbf{I}_i)) \notag \\
&\quad + \frac{\lambda}{|\mathcal{D}_U|} \sum_{j=1}^{|\mathcal{D}_U|} D_{MSE}\left(f_{{\theta}'}(x_{j},\eta') \| f_{\theta}(\mathbf{I}_j,\eta)\right)
\end{align}
The first term is the supervised loss, which is the difference between the student network prediction $f_{\theta}(x_{i})$ and label $y_{i}$ for labeled data. The second term is the consistency loss between predictions of the student with input noise $\eta$ and teacher networks for unlabeled data with noise $\eta'$. $\lambda$ is a weighting coefficient that controls the relative importance of the consistency loss compared to the supervised loss. 

Research \citep{yu2019uncertainty,burton2020semi} found that predictions from the teacher model in MT can be unreliable and noisy. Therefore, Monte Carlo Dropout was introduced to evaluate prediction uncertainty, allowing for the preservation of only those with low uncertainty when calculating the consistency loss. \citep{zeng2021reciprocal} argues that EMA focuses on weighting the student’s parameters during the training process rather than purposefully updating the parameters through an automatic evaluation strategy; thus, a meta-learning concept is utilized for the teacher model to generate better pseudo labels by observing how these labels affect the student' prediction. 

In recent years, dual-teacher frameworks have emerged as a promising direction for enhancing model diversity and robustness. To address the cardiac segmentation task, \cite{xiao2022efficient} proposed a dual-teacher model, where models A and B (a CNN and a ViT) capture both local and global information from the images to guide the student model's learning. Dual Consistency Mean Teacher (DC-MT) \citep{huo2022automatic} jointly regularizes both the output classification results and the attention masks (Class Activation Map, highlight the specific regions relevant for the classification task) for teacher and student model, to make the process more explainable. 

Unlike previous approaches that focus on sample-level prediction consistency in MT, \citep{liu2020semi} believes that the relationships among samples could facilitate learning useful semantic information from unlabeled data. \citep{liu2020semi} promotes consistency among different samples, emphasizing that samples with high similarity should remain closely related even after perturbations are introduced for classification tasks. Similarly, Graph Temporal Ensembling (GTE) \citep{shi2020graph} learns the relation between data. For labeled data, the method exploits connections by using a graph to map all labeled samples of the same class into a single cluster, thereby enhancing intra-class feature consistency. For unlabeled data, it guides the EMA-generated feature representations to move closer to their predicted cluster centers, leveraging semantic information implicitly present in the data. 




\subsubsection{Graph-based Methods}
Graph-based methods represent data relationships using graph structures, where labeled and unlabeled data are treated as nodes, and labels are propagated from labeled nodes to unlabeled ones based on their connectivity, following the smoothness assumption that neighboring data points should share the same class label \citep{chen2022semiun}. The convexity of graph-based Semi-SL simplifies optimization, making it easier to find local solutions, while its scalability enables efficient handling of large-scale datasets \citep{chong2020graph}. 

\cite{ganaye2018semi,ganaye2019removing} introduce a graph-based Semi-SL method, which is effective in segmenting brain MRI images. It uses labeled data to train a segmentation network and constructs an adjacency graph of anatomical structures. The method evaluates segmentation results on unlabeled images against the prior anatomical knowledge encoded in the adjacency graph, and the segmentation model is updated to make its predictions more consistent with the anatomical structure.



\subsubsection{Generative Methods} 
The generative methods utilize generative models such as AEs or GANs to learn feature representations from unlabeled data in Semi-SL, including adversarial and reconstruction learning. 

\textit{Adversarial Learning} can be generally categorized into two common strategies: learning the distribution of unlabeled data through adversarial training, where the discriminator guides the generator to produce realistic samples and exploits unlabeled data for better representation learning \citep{salimans2016improved}; or augment the training set by generating synthetic labeled samples that improve the performance of the semi-supervised model \citep{odena2017conditional}. 

For the first category, the discriminator finally outputs K+1 classes, where K is the real data categories for the dataset, and 1 bit indicates whether it is a generated image or real image (labeled and unlabeled data) \citep{salimans2016improved}, to facilitate learning the real image's distribution from the overall real data, including those unlabeled ones. The overall loss involves supervised and unsupervised loss, where supervised loss minimizes the classification error for labeled data by ensuring predictions align with the ground truth: 
\begin{equation}
L_{supervised} = -E_{\mathbf{I}_i, y \sim p_{data}(\mathbf{I}_i, y)} \log p_{model}(y | \mathbf{I}_i, y < K+1)
\end{equation}
For unsupervised loss, the first term encourages the model to classify labeled and unlabeled data as "real". The second term encourages the model to classify generated data as "fake". 
\begin{equation}
\begin{split}
L_{unsupervised} = & -E_{\mathbf{I} \sim p_{data}(\mathbf{I})} 
\log \big(1 - p_{model}(y = K+1 | \mathbf{I})\big) \\
& - E_{\mathbf{I} \sim G} \log p_{model}(y = K+1 | \mathbf{I})
\end{split}
\end{equation}

However, \cite{liu2019joint} points out that previous work \citep{salimans2016improved} has encountered challenges due to the competing roles of identifying fake samples and predicting class labels via a single discriminator. Therefore, the authors introduce a segmentation network for the main task, while the discriminator just determines whether the image-label pairs originate from the empirical joint distribution. A similar method is applied for Breast Ultrasound Mass Classification \citep{pang2021semi} with generated images for augmentation to learn more diverse data distributions. Papers \citep{nie2018asdnet,han2020semi} further get rid of the generator. To boost the segmentation task, \citep{nie2018asdnet} employs a network for segmentation and a discriminator that assesses the credibility of each region in the label maps produced by the network, to produce more accurate segmentation masks. Following this, \cite{han2020semi} notes that the confidence map presented by the discriminator may contain noise; therefore, a discriminator is specially designed to evaluate both target geometric and pixel-level information by using dual-attention from the predicted segmentation maps and original images. This design enhances the distinction between lesion areas and the background, thereby improving the accuracy of segmentation. However, this approach has become less prevalent, as balancing target learning with adversarial alignment remains nontrivial. The use of multiple networks or specialized discriminators often introduces additional complexity and increases the burden of hyperparameter tuning.

As another method, GANs \citep{goodfellow2014generative} are renowned for image generation capabilities, offering opportunities for generating a substantial quantity of synthetic medical images. These can be utilized for data augmentation purposes \citep{gupta2019generative, kamran2021vtgan}, which has become a hot topic these years. Despite their potential, generating anatomically plausible synthetic data remains a critical challenge. \citep{xing2023less} notes that generative models often fail to preserve human anatomical structures, leading to unrealistic features that weaken data augmentation. To address this, they propose a superpixel-based algorithm to extract structural guidance from unlabeled data, which conditions a conditional GAN for synthesizing images and annotations in a semi-supervised framework, enhancing learning with limited annotations. Similarly, \citep{xu2019semi} noted that conventional tumor generation methods often manually or randomly define tumor sizes and positions, which disrupt image priors and increase the risk of false positives in subsequent processing. Therefore, they employed CycleGAN (designed for image-to-image translation tasks, where the goal is to learn the transformation between an input image and an output image without paired examples) \citep{zhu2017unpaired} with attention modules to enhance focus on tumor regions during image translation. \citep{kugelman2023enhanced} harnesses StyleGAN \citep{karras2019style} to merge style information from unlabeled data with content from labeled data. This style fusion approach ensures that the newly generated labeled samples are not only diverse in style but also realistic, enhancing the model’s generalization capabilities to new, unseen data. 

\textit{Reconstruction Learning} are frequently used as an auxiliary task for segmentation, given the shared focus on pixel-level representation learning \citep{chen2022generative, chen2019multi}. Notably, \cite{chen2022generative} tackles minute cerebrovascular segmentation, which takes the segmentation model’s output as input and reconstructs the original data to improve the texture representation of the model. Conversely, \citep{chen2019multi} deviates from reconstructing the original input and instead trains an AE to reconstruct the foreground and background regions separately, which are divided by pseudo-segmentation labels. This approach encourages the model to learn segmentation-relevant features from unlabeled images. \citep{xie2019semi} recognizes that while an AE is a generative model optimal for learning representations suitable for reconstruction, it may not be ideal for the main task. Thus, they trained the model with reconstruction and classification tasks using both unlabeled and labeled data in a non-parameter-sharing manner for most layers, while employing learnable transition layers (just a few layers of the model) that enable the adaptation of the image representation ability learned by the reconstruction network to a lung nodule classification network. 

\subsubsection{Hybrid Methods}
\label{Holistic Methods}
In recent years, hybrid methods have been proposed, which combine multiple ideas, such as consistency regularization, data generation, and pseudo-labeling. Representative examples include Mixup \cite{zhang2017mixup}, MixMatch \citep{berthelot2019mixmatch}, and FixMatch \cite{sohn2020fixmatch}. These strategies reflect a growing emphasis on exploiting both labeled and unlabeled data through more flexible augmentations. This not only helps the model learn from unlabeled data but also expands the diversity of the training dataset as an augmentation, leading to improved generalization. 

Taking Mixup \citep{zhang2017mixup} as an example, in Semi-SL, this method performs data augmentation by linearly interpolating labeled and unlabeled data and their corresponding labels (given or predicted). Unlike traditional augmentations such as rotation or flipping, Mixup creates synthetic training samples by blending two images $\mathbf{I}_{1}$, $\mathbf{I}_{2}$ together to form a new image ${\mathbf{I}}'$ with label ${y}'$ by the following process: 
\begin{equation}
{\lambda }'=max(\lambda,1-\lambda) \ \ where \ \lambda \sim Beta(\alpha ,\alpha ) 
\end{equation}
\begin{equation}
{x}'={\lambda }'e_{l}(\mathbf{I}_{1})+(1-{\lambda}')\cdot e_{l}(\mathbf{I}_{2}) 
\end{equation}
\begin{equation}
{y}'={\lambda }'\cdot y_{1}+(1-{\lambda }')\cdot y_{2}
\end{equation}

Then, the generated mixed data are used to train the network, encouraging the model to learn smoother decision boundaries between classes, thereby improving its generalization ability.

Mixup can be used at an image or feature level to take unlabeled data representation into learning. \citep{gyawali2020semi} continuously updated pseudo labels from an average of predictions of augmented samples for each unlabeled data point. Then performs mixing of pairs of labeled and unlabeled data, both in the input and latent space, along with their corresponding labels. This fills the gaps between input samples, thus training the deep network more effectively. To tackle the challenges of 3D medical image detection, \citep{wang2020focalmix} adapted the MixUp data augmentation strategy at both the image and object levels to accommodate the medical image detection tasks. Overall, these methods, as a hot topic, highlight a burgeoning trend toward more sophisticated data mixing protocols that address unique medical imaging challenges (e.g., multi-organ segmentation, and 3D object detection).

More recently, mixing strategies have been increasingly combined with other learning paradigms, becoming a popular research direction. MagicNet \citep{chen2023magicnet}, designed for multi-organ CT image segmentation, enhances the model’s ability to learn about the varying sizes and positions of different organs by mixing image patches cross or within the image, to capture the intrinsic structure and relationships of the images, aiding the MT model in learning. In \cite{yuan2023semi}, a prototype-based classifier that utilizes characteristic features of each class guiding teacher model learning. To train the student, a part is cut from a labeled image (foreground) and pasted onto an unlabeled image (background), and vice versa to create new mixed images with teacher predictions \citep{bai2023bidirectional}. 

\begin{table*}[]
\centering
\caption{Overview of Semi-supervised Learning Strategies in Medical Imaging Analysis (Part 1).}
\label{tab:SSL}
\scalebox{0.88}{
\begin{tabular}{llllll}
\hline
Reference & Category & Sub-category & Body & Modality & Task \\ \hline
\cite{xu2024expectation} & Pseudo Labeling & Self-training & Brain/Vessel/Prostate & CT/MRI & Cls \\
\cite{han2022effective} & Pseudo Labeling & Self-training & Liver & CT & Seg \\
\cite{wu2022minimizing} & Pseudo Labeling & Self-training & Cardiac/Eye & MRI/OCT & Seg \\
\cite{huo2021atso} & Pseudo Labeling & Co-training & Pancre & CT & Seg \\
\cite{zhu2021semi} & Pseudo Labeling & Co-training & Cardiac & MRI/CT & Seg \\
\cite{wu2023compete} & Pseudo Labeling & Co-training & Cardiac/Pancre/Colon & MRI/CT/CT & Seg \\
\cite{bortsova2019semi} & Pseudo Labeling & Co-training & Chest & X-ray & Seg \\
\cite{luo2022semi} & Pseudo Labeling & Co-training & Cardiac & MRI & Seg \\
\cite{shen2023co} & Pseudo Labeling & Co-training & Skin/Cardiac/Colon & Derm/MRI/CT & Seg \\
\cite{su2024mutual} & Pseudo Labeling & Co-training & Cardiac/Pancre/brain & MRI/CT & Seg \\
\cite{xie2023deep} & Pseudo Labeling & Co-training & Cardiac & MRI & Seg \\
\cite{xu2019deepatlas} & Consistency Regularization & Feature Consistency & Brain/knee & MRI & Seg/Reg \\
\cite{berenguer2024semi} & Consistency Regularization & Feature Consistency & Chest/Skin & X-ray/CT/Derm & Cls \\
\cite{wu2022mutual} & Consistency Regularization & Feature Consistency & Cardiac/Pancre & MRI/CT & Seg \\
\cite{peiris2021duo} & Consistency Regularization & Feature Consistency & Nuclei/Heart/Spleen & Histo/MRI/CT & Seg \\
\cite{luo2021efficient} & Consistency Regularization & Feature Consistency & NPC dataset & MR & Seg \\
\cite{luo2022semi1} & Consistency Regularization & Feature Consistency & Brain/Pancre/Pharynx & MRI & Seg \\
\cite{xia20203d} & Consistency Regularization & Feature Consistency & Pancre & CT & Seg \\
\cite{liu2021federated} & Consistency Regularization & Feature Consistency & Brain/Skin & CT/Derm & Cls \\
\cite{chen2022mass} & Consistency Regularization & Feature Consistency & Abd & CT & Seg \\
\cite{luo2021semi} & Consistency Regularization & Task Consistency & Atrium/Pancre & MRI/CT & Seg \\
\cite{chen2022semi} & Consistency Regularization & Task Consistency & Brain/Cardiac & MRI & Seg \\
\cite{cao2020uncertainty} & Consistency Regularization & Self-ensembling & Skin & Derm & Seg \\
\cite{yu2019uncertainty} & Consistency Regularization & Mean Teacher & Cardiac & MRI & Seg \\
\cite{liu2020semi} & Consistency Regularization & Mean Teacher & Chest/Skin & X-ray/Derm & Cls \\
\cite{burton2020semi} & Consistency Regularization & Mean Teacher & Knee & MRI & Seg \\
\cite{shi2020graph} & Consistency Regularization & Mean Teacher & Cell & Histo & Cls \\
\cite{yuan2023semi} & Consistency Regularization & Mean Teacher & Cardiac & MRI & Seg \\
\cite{bai2023bidirectional} & Consistency Regularization & Mean Teacher & Cardiac/Pancre & MRI & Seg \\
\cite{wang2023mcf} & Consistency Regularization & Mean Teacher & Cardiac/Pancre & MRI/CT & Seg \\
\cite{zeng2021reciprocal} & Consistency Regularization & Mean Teacher & Cardiac/Pancre & MRI/CT & Seg \\
\cite{xiao2022efficient} & Consistency Regularization & Mean Teacher & Cardiac & MRI & Seg \\
\cite{huo2022automatic} & Consistency Regularization & Mean Teacher & Knee & MRI & Cls \\
\cite{ganaye2018semi} & Graph-based & / & Brain & MRI & Seg \\
\cite{ganaye2019removing} & Graph-based & / & Brain & MRI & Seg \\
\cite{liu2019joint} & Generative Learning & Adversarial Learning & Eye & OCT & Seg \\
\cite{nie2018asdnet} & Generative Learning & Adversarial Learning & Prostate & MRI & Seg \\
\cite{han2020semi} & Generative Learning & Adversarial Learning & Breast & US & Seg \\
\cite{pang2021semi} & Generative Learning & Adversarial Learning & Breast & US & Cls \\
\cite{gupta2019generative} & Generative Learning & Augmentation & Bone & X-ray & Cls \\
\cite{kamran2021vtgan} & Generative Learning & Augmentation & Eye & OCT & Cls \\
\cite{kugelman2023enhanced} & Generative Learning & Augmentation & Eye & OCT & Seg \\
\cite{xing2023less} & Generative Learning & Adversarial Learning & Chest & CT & Synthesis \\
\cite{xu2019semi} & Generative Learning & Adversarial Learning & Brain & MRI & Cls \\
\cite{chen2022generative} & Generative Learning & Reconstruction & Brain & MRI & Seg \\
\cite{chen2019multi} & Generative Learning & Reconstruction & Brain & MRI & Seg \\
\cite{xie2019semi} & Generative Learning & Reconstruction & Chest & CT & Cls \\
\cite{gyawali2020semi} & Hybrid & / & Chest/Skin & X-ray/Derm & Cls \\
\cite{wang2020focalmix} & Hybrid & / & Chest/Thoracic & X-ray/CT & Det \\
\cite{shin2018joint} & Pseudo Labeling (ISL) & Self-training & Breast & US & Det/Cls \\
\cite{zhou2019collaborative} & Pseudo Labeling (ISL) & Self-training & Eye & OCT & Seg \\
\cite{mlynarski2019deep} & Pseudo Labeling (ISL) & Self-training & Brain & MRI & Seg \\
\cite{hu2021semi} & Pseudo Labeling (UL) & Self-training & Brain/cardiac & MRI/CT & Seg \\
\cite{chaitanya2020contrastive} & Pseudo Labeling (UL) & Self-training & cardiac & MRI/ & Seg \\
\cite{feng2018deep} & Pseudo Labeling (UL) & Self-training & Cell & histo & Cls \\
\cite{shi2021semi} & Pseudo Labeling (TL) & Self-training & Chest & CT & Cls \\
\cite{liu2023segmentation} & Pseudo Labeling (ISL) & Self-training & Cardiac & MRI & Seg \\
\cite{marini2021semi} & Pseudo Labeling (ISL/TL) & Self-training & Tissue & Histo & Cls \\
\cite{miao2023sc} & Pseudo Labeling (UL) & Self-training & Cardiac/Pancre & MRI/CT & Seg \\
\cite{wang2021learning} & Pseudo Labeling (ISL) & Self-training & Pancre & CT & Seg \\
\cite{li2022deep} & Pseudo Labeling (ISL) & Self-training & Breast & US & Seg \\
\cite{zhang2022boostmis} & Consistency Regularization (AL) & Feature Consistency & Respiratory & MRI & Cls \\
\cite{gao2022segmentation} & Consistency Regularization (ISL) & Mean Teacher & Cardiac & MRI & Seg \\
\hline
\end{tabular}}
\end{table*}

\begin{table*}[]
\centering
\caption*{\textbf{Table 3 (continued).} Overview of Semi-supervised Learning Strategies in Medical Imaging Analysis (Part 2).}
\scalebox{0.88}{
\begin{tabular}{llllll}
\hline
Reference & Category & Sub-category & Body & Modality & Task \\ \hline

\cite{zhou2023recist} & Consistency Regularization (ISL) & Mean Teacher & Abd & CT & Seg \\
\cite{chen2023magicnet} & Consistency Regularization (UL) & Mean Teacher & Abd & CT & Seg \\
\cite{zhao2023rcps} & Consistency Regularization (UL) & DAC & Cardiac/Pancre & MRI/CT & Seg \\ \hline
\end{tabular}}
\end{table*}

\subsubsection{Integration with other techniques}
\label{SSLIntegration with other techniques}


Semi-SL has shown promise in leveraging unlabeled medical data, but often suffers from the unreliability of pseudo-labels. This can lead to models learning imprecise feature representations, which adversely affect diagnostic outcomes. Therefore, Semi-SL can be integrated with other paradigms such as AL, inexact supervision, unsupervised methods, or transfer learning, to optimize its performance. 

\textit{AL} can be integrated to select and learn those informative data that predict confidence below the predefined threshold to offer more precise labeled data for training \citep{zhang2022boostmis}. This with Semi-SL creates a synergistic effect where both methods mutually benefit each other. 

Instead of requiring more accurate human-labeled data, \textit{Inexact supervision} demonstrates potential in helping Semi-SL, which can further expand the utilization of data with coarse annotations, aiding in the generation of more accurate pseudo labels. One approach involves sharing a single model to jointly learn from different levels of annotations. For example, \cite{shin2018joint} addresses the risk of overfitting with small, strongly annotated datasets in Semi-SL by incorporating inexact labels to improve localization and classification of masses within the same localization model. Similarly, given the correlation between severity grading in Diabetic Retinopathy (DR) and lesion characteristics, \cite{zhou2019collaborative} pre-trains a segmentation model using limited pixel-level annotations and then generates pseudo lesion masks for images with only disease grade labels using an attention mechanism. To segment cancerous lesions, which are typically smaller than the overall pancreatic region, \cite{wang2021learning} incorporates attention mechanisms into a multiple instance learning (MIL) framework, an inexact supervised method which will be discussed in Section \ref{Multiple-Instance Learning}. \cite{li2022deep} utilized CAM for identifying breast tumors, with the breast anatomical structures initially segmented and used as constraints, which are guided by full labels. 

Another approach utilizes dual-branch architectures to handle inexact and full labels separately, mitigating the learning difficulty caused by training a single model on heterogeneous supervision, while enabling mutual guidance through weight sharing. For instance, \cite{mlynarski2019deep} extends segmentation networks with an additional branch for image-level classification, leveraging inexact labels while mitigating the risk of learning irrelevant features. Although sharing a feature space can be beneficial, it is noted that this method might lead to suboptimal feature representations for the mainstream task \citep{liu2023segmentation}. \cite{marini2021semi} utilized a teacher-student model, where a high-capacity teacher model was trained on all strong labels and used to label a set of pseudo-label samples, which were then used to train a student model. The student model was further fine-tuned with strongly-annotated data to improve its performance. Above that, \citep{liu2023segmentation} proposed KL divergence controlling the distance between teacher and student brand, for further guidance. The third approach generates pseudo-labels from inexact annotations like RECIST markers \citep{beaumont2019radiology}, which indicate tumor size for guiding segmentation. \citep{zhou2023recist} further refines lesion boundaries using a structure-aware graph built from RECIST annotations.

Beyond inexact annotations, \textit{UL}, particularly Self-SL, can further assist feature learning from unlabeled data. One way is to use unlabeled data for pretraining then finetuned by limited labeled data. \cite{hu2021semi} employs contrastive learning (a common Self-SL approach in Section \ref{Discriminative Model}) for the pre-training stage by introducing a new loss function to cluster pixels with the same label closer together in the embedding space, enhancing feature discrimination. Then limited labeled data are used to finetune the model for segmentation tasks. Likewise, \cite{chaitanya2020contrastive} proposed a framework combining global and local contrastive losses for training 3D medical image segmentation models. The global loss aligns features from two augmented versions of the same 3D volume, while the local loss focuses on patch-level features, treating features at the same location across augmentations as positive pairs and others as negatives. \citep{feng2018deep} uses AE for reconstruction with an additional constrain that minimizes the distance between patches with high similarity in both the input space and the hidden space to preserve the geometric structure of the data as a pre-trained model for further labeled data fintuning. Contrastive loss can further strengthen the representation learning during the training process \citep{chaitanya2023local,miao2023sc,zhao2023rcps}. For example, in \cite{chaitanya2023local}, local contrastive loss is created based on pseudo-labels, where pixels from the same class should be positives. Then both local contrastive loss and segmentation supervised loss are optimized together. 

\textit{Transfer learning} can assist Semi-SL by incorporating more diverse data when labeled samples are limited \citep{liu2021semi}. \cite{liu2021semi} employs a disentanglement technique to separate domain-specific and domain-invariant representations. The domain-specific parts are used to reconstruct input images, while domain-invariant features support downstream tasks. This separation enables learning generalizable features across domains, improving Semi-SL under limited labels. Given the limited number of pathology samples, \cite{shi2021semi} utilized a pre-trained network to classify lung nodules and similar tissues as source domain. The model predicts labels for unlabeled nodules and selectively incorporates them into the training dataset, refining its performance.


\paragraph{Summary of Semi-supervised learning} Table \ref{tab:SSL} highlights representative studies on Semi-SL. Among the various approaches, pseudo-labeling methods and consistency regularization have emerged as the most widely adopted strategies, both historically and in recent work, due to their conceptual simplicity and ease of implementation. Pseudo-labeling enables models to exploit unlabeled data by assigning predicted labels as supervisory signals. This exposes the model to a broader data distribution and promotes generalization, even when the pseudo-labels are imperfect. However, when labeled data are extremely scarce or the data distribution exhibits a strong domain shift, incorrect pseudo-labels can accumulate and reinforce erroneous patterns, ultimately degrading performance. In practice, consistency-based methods, by enforcing prediction invariance under input perturbations, model perturbations, or task variations, tend to exhibit greater robustness, particularly in heterogeneous or noisy clinical datasets. 

In recent years, the integration of Semi-SL with other learning paradigms has gained increasing attention. Notably, self-supervised learning has emerged as a complementary component due to its compatibility and ease of incorporation, often through pretraining or simple loss-based constraints, which is expected to persist in the future. 



\section{Inexact Supervision}
\label{sec: Inexact Supervision}
\begin{figure}[tbp]
\centering
\includegraphics[width=0.33\textwidth]{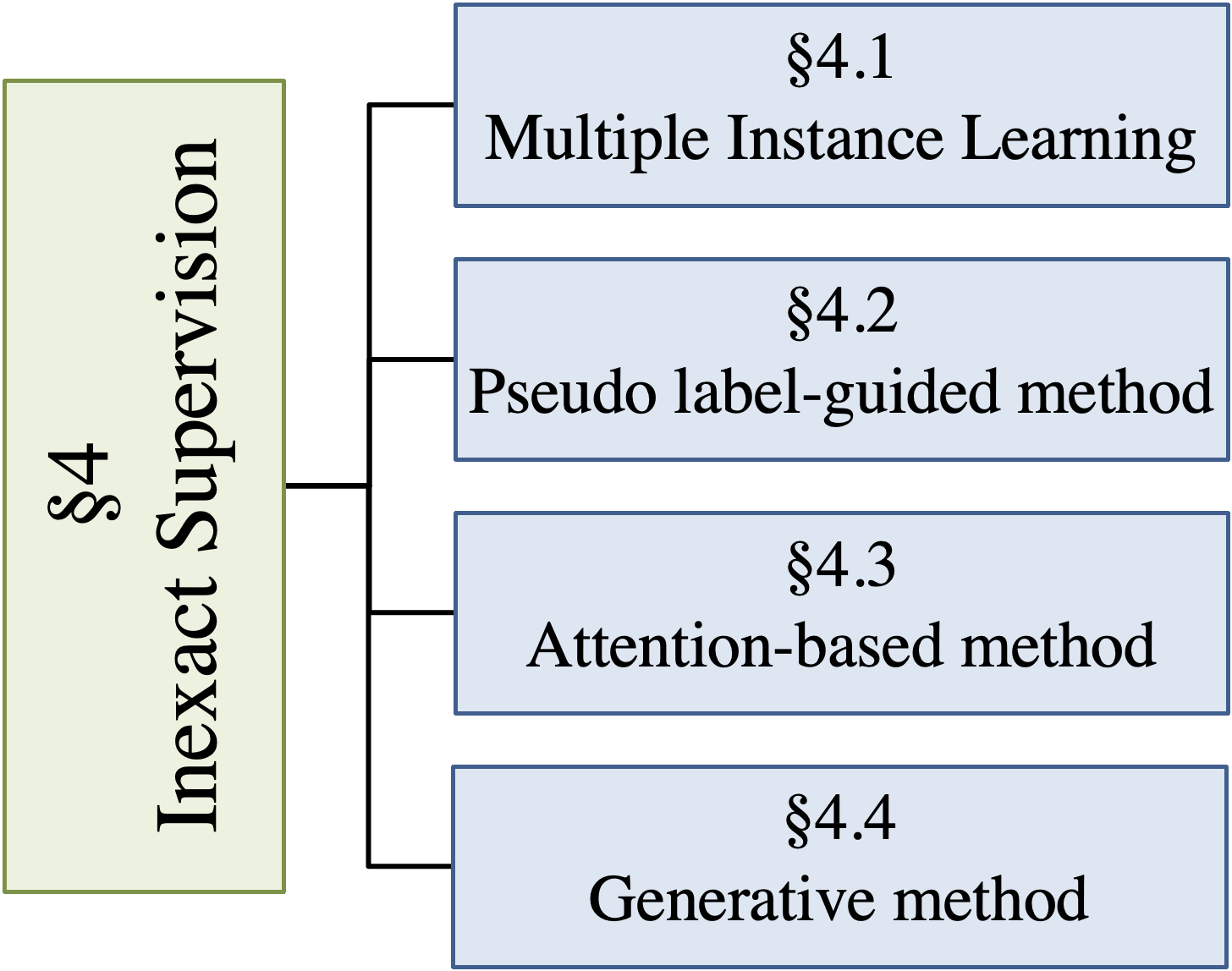}
\caption{The taxonomy of Inexact-supervised Learning.}
\label{fig:inexact_tree}
\end{figure}

Inexact supervision refers to learning from inexact or coarse labels, where each data is partially supervised and lacks precise annotations \citep{zhou2018brief}. For example, a chest X-ray dataset may only include labels indicating the presence or absence of a condition (e.g., Covid‐19) without detailed region annotations, requiring the model to infer fine-grained details (e.g., lesion boundaries). The formal definition is in Definition \ref{def: Inexact Supervision}, and its taxonomy is shown in Fig. \ref{fig:inexact_tree}, where methods are categorized into multiple-instance learning, pseudo label-guided, attention-based, and generative approaches.

\subsection{Multiple-Instance Learning}
\label{Multiple-Instance Learning}

Multiple-Instance Learning (MIL) \citep{dietterich1997solving} addresses inexact labeled data by training models to make predictions based on sets of instances (e.g., image patches), where only set-level labels (e.g., disease presence) are provided, without instance-level annotations \citep{zhou2018brief}. 

Let $\mathcal{D}_L^* = \{ (\mathcal{I}_i, y_i) \}_{i=1}^{|\mathcal{D}_L^*|}$ denote the training dataset, where each $\mathcal{I}_i = \{ \mathbf{I}_{i,j} \}_{j=1}^{|\mathcal{I}_i|}$ is a bag composed of multiple image patches. A bag $\mathcal{I}_i$ is labeled as positive if at least one instance $\mathbf{I}_{i,j} \in \mathcal{I}_i$ is positive; otherwise, it is labeled as negative. The objective of MIL is to predict the labels of unseen bags.

Note that up to now, most MIL-related works are fully supervised to enhance accuracy \citep{dov2021weakly,su2022attention2majority}, while our focus is primarily on methods that address finer-grained tasks, aiming to reduce labeling costs. For example, \cite{schwab2020localization} performs lesion localization using only image-level labels. Image patches are processed by a classification model to obtain patch class scores. Max-pooling aggregates these scores for slide-level classification, while the most discriminative patches are identified for localization.

MIL can be used for detection tasks under inexact supervision, but it is unsuitable for segmentation, particularly when the target structures are small or have irregular boundaries. This limitation arises from its reliance on instance-level predictions over predefined patches, which may fail to capture fine-grained or boundary-level details.

\subsection{Pseudo label-guided method}
Directly generating pseudo-labels at the desired level is an intuitive method under inexact supervision. This process closely resembles the pseudo labeling approach in Semi-SL (Section~\ref{Pseudo Labeling}), except that inexact labels are employed across all samples. For instance, \citet{li2019weakly} expand a centroid point label (given inexact supervision) into a concentric circle design, where the inner circle indicates the mitotic region and the surrounding ring handles uncertain pixels. By excluding the loss on the ring region, the model avoids noise and focuses more effectively on mitotic segmentation. \citet{valvano2021learning} employs a GAN, in which the generator predicts realistic segmentation masks while the discriminator, guided by scribble constraints, differentiates generated masks from unpaired real masks.  \citep{gao2022segmentation} proposes a mean-teacher framework where the student model is weakly supervised by scribbles and a Geodesic distance map \citep{wang2018deepigeos} derived from scribbles and aligns predictions with the teacher model along horizontal and vertical axes for segmentation task. 

\subsection{Attention-based method}
\begin{figure}[tbp]
\centering
\includegraphics[width=0.9\linewidth]{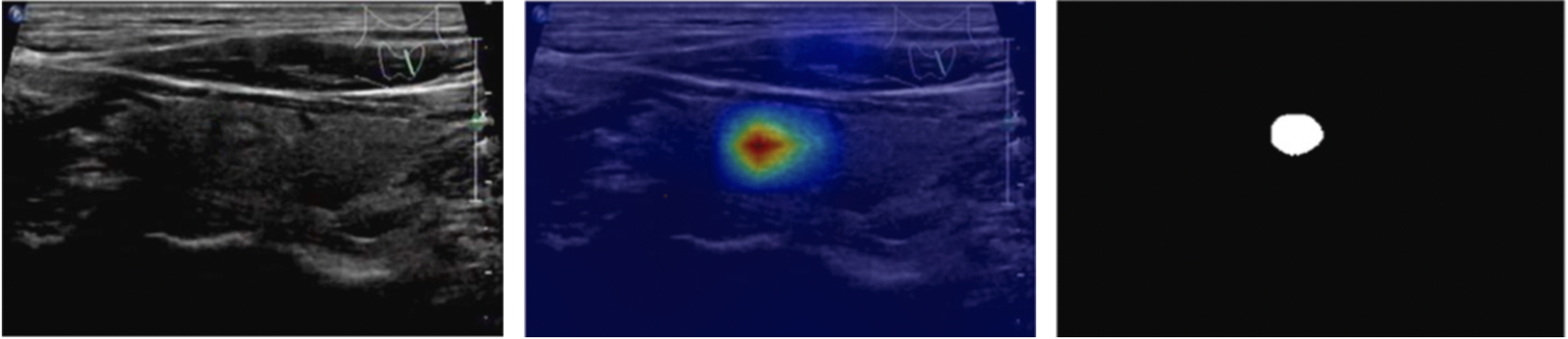}
\caption{An example of CAM of a nodule in thyroid ultrasound images \citep{yu2022adaptive}, where the left is the input image, the middle one is the visualization of the model CAM, and the last image denotes ground truth.}
\label{fig:CAM}
\end{figure}

Class Activation Map (CAM) \citep{zhou2016learning}, also known as a class heat map or saliency map, is a visual representation of equal size to the original input. CAM indicates the relative contribution of the corresponding region in the input image to the predicted output. Higher scores correspond to regions with a stronger response and greater influence on the network's decision, shown in Fig. \ref{fig:CAM}. The CAM is typically generated by fusing the channels of the feature map from the last convolutional layer, weighted by their contributions to the output class.


\begin{table*}[]
\centering
\caption{Overview papers of Inexact-supervised Learning in Medical Imaging Analysis.}
\label{tab:inexact}
\scalebox{0.88}{
\begin{tabular}{llllll}
\hline
\multicolumn{1}{l}{Reference} & Category & Body & Modality & Label Given & \multicolumn{1}{l}{Task} \\ \hline
\cite{dov2021weakly} & MIL & Tissue & Histo & Image-Level & ROI Cls \\
\cite{su2022attention2majority} & MIL & Tissue & Histo & Image-Level & ROI Cls \\
\cite{li2019weakly} & PL-guided & Cell & Histo & Centroid Point & Det \\
\cite{valvano2021learning} & PL-guided  & Cardiac/Abd & MRI & Scribble & Seg \\
\cite{wang2020weakly} & Attention-based  & Chest & CT & Image-Level & Cls/Det \\
\cite{dubost2020weakly} & Attention-based  & Brain & MRI & Image-Level & Det \\
\cite{taghanaki2019infomask} & Attention-based  & Chest & X-Ray & Image-Level & Det \\
\cite{seebock2018unsupervised} & Generative  & Eye & OCT & Image-Level & Det \\
\cite{pinaya2022unsupervised} & Generative  & Head & CT & Image-Level & Det \\
\cite{zhao2021anomaly} & Generative  & Chest/Eye & X-ray/OCT & Image-Level & Det \\
\cite{bercea2023reversing} & Generative  & Brain & MRI & Image-Level & Det \\
\cite{pinaya2022fast} & Generative  & Head/brain & CT/MRI & Image-Level & Det \\
\cite{kuang2023cluster} & Attention-based (UL) & Brain/Prostate & MRI & Image-Level & Seg \\
\cite{liang2021contrastive} & PL-guided  (Semi-SL/UL) & Chest & X-Ray & Report & Cls \\
\cite{schirris2022deepsmile} & MIL (UL) & Tissue & Histo & Image-Level & ROI Cls \\ \hline
\end{tabular}}
\end{table*}

\cite{wang2020weakly} noted that, specifically for Covid-19, the boundaries can be indistinct. Given image-level labels, a pre-trained model is first employed to segment the lung regions from 3D CT scans to isolate areas of interest that may contain lesions. Subsequently, CAM is used for Covid-19 lesion segmentation. In another way, \cite{dubost2020weakly} developed a method to detect small target lesions by calculating attention maps from the final feature maps of a segmentation network at full input resolution without interpolation, allowing small lesions to appear prominently in the attention maps. The segmentation network optimizes solely through a global regression objective, which is the count of brain lesions. Given that attention maps can be noisy, leading to erroneously highlighted regions. Authors \citep{taghanaki2019infomask} introduced a learned spatial masking mechanism designed to filter out irrelevant backgrounds from attention maps for chest disease localization. This is achieved by minimizing the mutual information between the representation and the input while maximizing the information shared between the masked representation and class labels. 


\subsection{Generative method}
Generative models, especially AEs learn key features by reconstructing input data. They are widely used for anomaly detection \citep{seebock2018unsupervised}, typically requiring image-level labels (e.g., healthy or not). By just learning the distribution of healthy data, the model reconstructs inputs accordingly. For anomalous samples, since the model has only been trained on healthy data, the model tends to generate a "healthy" version of the input. The difference between the original input and the reconstructed image highlights the anomaly region. Building on this, to solve unsupervised brain imaging 3D anomaly detection, \citep{pinaya2022unsupervised} feeds the "healthy" latent tokens produced by a VAE into a transformer, which autoregressively predicts the next latent element, capturing long-range structural dependencies and spatial relationships between different brain regions. During inference, elements with significant prediction errors are considered anomalies.

However, AE-based methods often produce blurry reconstructions. \cite{zhao2021anomaly} further uses GAN for reconstruction with two key constraints: healthy features should be tightly clustered, and the discriminator enforces reconstruction realism. This cross-space regulation ensures consistent and representative features, enabling effective anomaly detection in OCT and chest X-ray datasets. Instead of reconstructing the entire image, \cite{bercea2023reversing} uses a GAN to first generate a coarse anomaly label to identify abnormal regions, which are masked and refined using an inpainting network (preserve healthy tissues and produce pseudo-healthy in-painting in anomalous regions). The final anomaly map is obtained by comparing the pseudo-healthy image with the original input. Denoising Diffusion Probabilistic Models (DDPM), a generative model that produces high-quality samples by progressively denoising data, can also be applied in a similar manner \citep{pinaya2022fast}.

However, AE-based methods often produce blurry reconstructions. \cite{zhao2021anomaly} further applies GANs for reconstruction with two key constraints: healthy features should be tightly clustered, and the discriminator enforces reconstruction realism. This cross-space regulation ensures consistent, representative features, enabling effective anomaly detection in OCT and chest X-ray datasets. Instead of reconstructing the entire image, \cite{bercea2023reversing} uses a GAN to first generate a coarse anomaly label to identify abnormal regions, which are masked and refined using an inpainting network (preserving healthy tissues and producing pseudo-healthy completions in anomalous regions). The final anomaly map is obtained by comparing the pseudo-healthy image with the original. Denoising Diffusion Probabilistic Models (DDPM), a generative model that produces high-quality samples via progressive denoising, can also be used similarly \citep{pinaya2022fast}.

\subsection{Integration with other techniques}
While ISL reduces labeling costs by providing coarse labels, these labels often lack sufficient information, restricting the performance. Thus, many other scenarios can be integrated with inexact supervision to further boost performance. For example, Semi-SL, which leverages a small subset of fully annotated data, can enhance model performance (see Section~\ref{Semi-supervised Learning (Semi-SL)} for details).

\textit{UL} explores relationships among samples without any explicit guidance, offering additional constraints for inexact supervision. CAM-based approaches often lead to under- or over-segmentation due to loose constrain provided by inexact labels. Therefore, \cite{kuang2023cluster} uses the feature representations learned from the initially generated CAMs as input to a clustering algorithm, which segments the image pixels into different categories and generates pixel-wise clustering maps. These maps are subsequently utilized as pixel-level supervision to expand under-activated regions (addressing under-segmentation) and suppress irrelevant activations (addressing over-segmentation), ultimately resulting in more refined and accurate CAMs. Furthermore, Self-SL can build a pre-trained model that aids in general feature extraction, helping further processing \citep{schirris2022deepsmile}. Specifically, the authors use SimCLR \citep{chen2020simple} to strengthen the latent representations of augmented versions of the same tile to be similar while ensuring that representations of different tiles are distinct, thereby training feature extractors. Further, attention mechanisms focus on the tiles most influential for classification to aid MIL. 

\paragraph{Summary of Inexact-supervised learning} Compared with incomplete and unsupervised supervision, this field appears relatively underexplored. It is frequently applied to histopathology datasets, where each image can be extremely large in size, making precise annotations particularly labor-intensive. Two primary approaches dominate this space: multiple-instance learning (MIL) and attention-based methods, often implemented via class activation maps (CAM). MIL refines predictions by analyzing patches within an image, frequently improving fully supervised baselines. CAM-based techniques, on the other hand, provide intuitive visual explanations but often face issues of under- or over-segmentation. Recent works \citep{chikontwe2022weakly, silva2021weglenet} attempt to mitigate these inaccuracies, yet achieving reliable and precise pixel-level delineations remains challenging, especially given the fine-grained demands of clinical applications.

In future development, we anticipate inexact supervision will remain a relatively less focus compared to the more actively explored incomplete and unsupervised paradigms. However, the concept of leveraging coarse user interactions (e.g., click-based prompts) has recently proven successful in foundational models like Segment Anything Model (SAM) \citep{kirillov2023segment}, underscoring the potential of partial or approximate annotations for rapid model adaptation.

\section{Absent Supervision}
\label{sec: Absent Supervision}
In medical image analysis, labeled data, whether precise or coarse, can be scarce or unavailable due to factors like rare diseases, expensive annotation, or privacy concerns, necessitating absent supervision \citep{chen2022semiun}. We primarily focus on two learning paradigms: unsupervised learning (UL), which relies solely on unlabeled datasets, and transductive transfer learning (TTL), which utilizes labeled source data to help learn from target unlabeled data.

\subsection{Unsupervised Learning (UL)}
\label{Unsupervised Learning}
UL trains models to discover patterns, structures, or representations in data without using labeled information:
\begin{definition}[Unsupervised Learning]
     UL is considered as training on a dataset with only input instances and no corresponding labels, where $f(\cdot)$ learns exclusively from $\mathcal{D}_U$.
\end{definition}

\begin{figure}[tbp]
\centering
\includegraphics[width=0.43\textwidth]{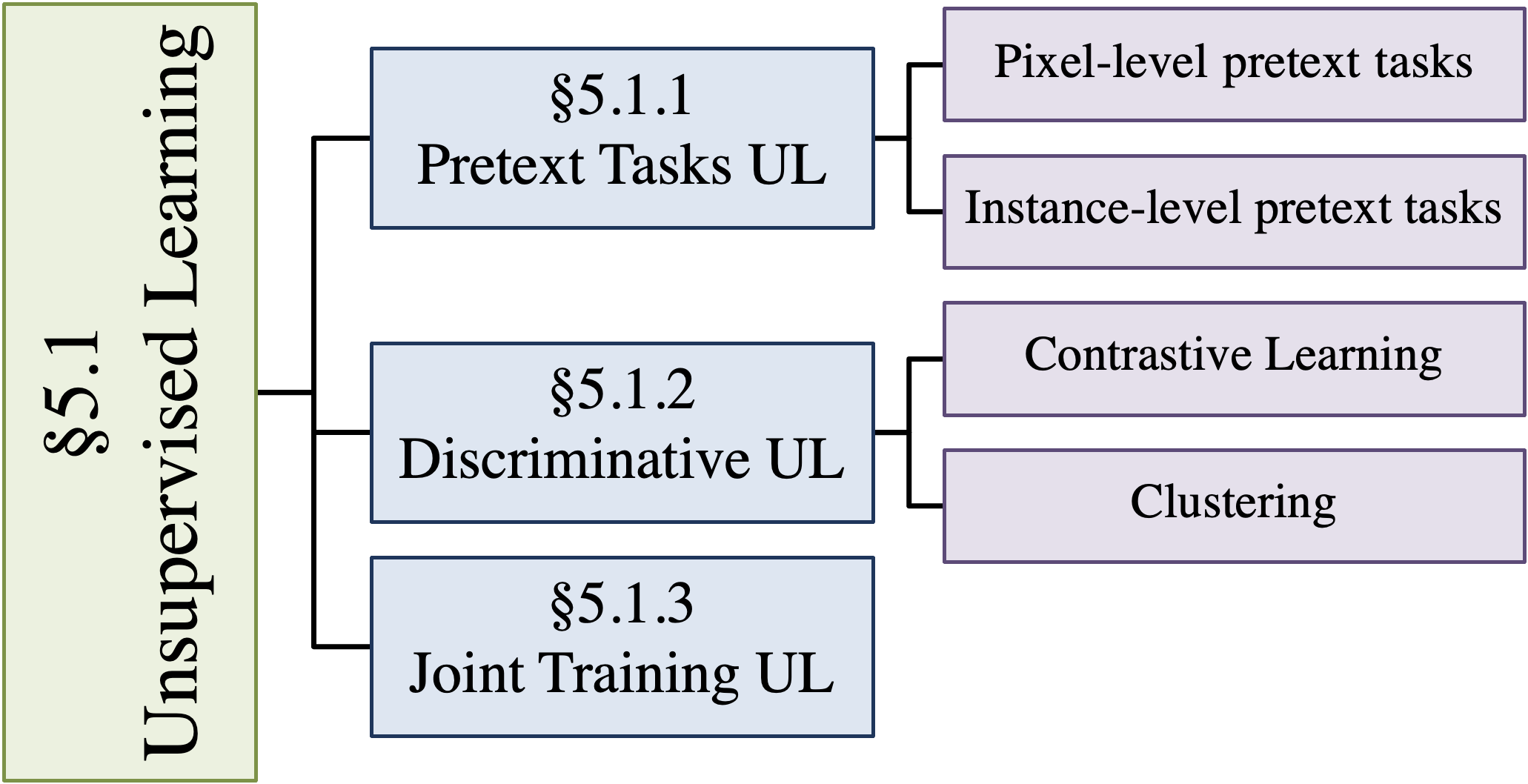}
\caption{The taxonomy of Unsupervised Learning.}
\label{fig:unsupervised_tree}
\end{figure}

By extracting semantic features from unlabeled data, UL improves the performance of downstream tasks such as classification or segmentation \citep{zhang2023dive}. Based on methodologies, UL can be categorized into three types: pretext tasks, using proxy objectives for representation learning; discriminative model, focusing on separating or clustering data; and joint training, simultaneously optimizing multiple learning objectives to enhance representation learning \citep{chen2022semiun}. Fig. \ref{fig:unsupervised_tree}
shows the overall structure of UL. 


\subsubsection{Pretext Tasks}
Pretext tasks are proxy tasks designed to teach models useful instance representations for specific training objectives through simpler problems without labels. These tasks help models extract semantic features to aid downstream tasks like classification or segmentation \citep{taleb20203d}. Based on the level of focus, pretext tasks can be classified into pixel-level tasks and instance-level tasks.

\textit{Pixel-level pretext tasks}, widely used in segmentation, aim to predict or reconstruct pixel values, enabling models to learn fine-grained spatial features \citep{chen2022semiun}. In medical imaging, common examples include masked image modeling, which reconstructs masked regions to capture contextual relationships and pathological features \citep{yan2023label,zhu2020rubik,xie2022s,jiang2022self}; jigsaw solving, which reassembles shuffled image patches to develop spatial understanding \cite{bai2019self}; and anatomy-oriented tasks, which identify or predict anatomical structures to enhance feature comprehension \citep{holmberg2020self}. Additionally, Scale-Aware Restoration (SAR) \citep{zhang2021sar,zhang2021self} boosts 3D tumor segmentation by teaching models to predict sub-volume scales and restore original volumes, fostering multi-scale feature learning.

\textit{Instance-level pretext tasks}, often used for classification, assign sparse semantic labels by generating diverse inputs (e.g., image rotation \cite{zhuang2019self}) and predicting transformations without annotations \citep{chen2022semiun}. To better align the pretext task with Covid-19 CT classification task, \cite{ewen2021targeted} proposed a proxy task involving flipping CT images horizontally and training the model to distinguish between original and flipped images, helping it learn distinguishing features closely related to the classification task.



\subsubsection{Discriminative UL}
\label{Discriminative Model}
Discriminative UL focuses on distinguishing or clustering data by learning boundaries or patterns within unlabeled data without explicitly modeling the underlying data distribution. They can be categorized into contrastive learning that maximizes differences between instances and clustering methods that group similar points \citep{chen2022semiun}.

\textit{Contrastive learning} (CL) trains models to learn meaningful instance representations that can be used for various downstream tasks (e.g., classification, segmentation, object detection) by pulling similar data points closer and pushing dissimilar ones apart in the feature space. CL achieves this by minimizing embedding distances for positive pairs (e.g., augmented views of the same image) and maximizing distances for negative pairs (e.g., different images) using a contrastive loss function \citep{li2021rotation,fernandez2022contrasting}. InfoNCE loss \citep{oord2018representation} is one of the most widely used contrastive losses as it leverages the noise contrastive estimation (NCE) principle \citep{gutmann2010noise} to maximize mutual information across views and achieve invariance. Formally, InfoNCE is expressed as:
\begin{equation}
    L_{infoNCE} = \sum_{\mathbf{I}_i \in \mathcal{D}_U} -\log \frac{\exp{(\frac{f(\mathbf{I}_i) \cdot f(\mathbf{I}'_i)}{\tau}})}{\sum_{\mathbf{I}_j \in \mathcal{D}_U}\exp(\frac{f(\mathbf{I}_i)\cdot f(\mathbf{I}_j)}{\tau})}
\end{equation}
where $f(\cdot)$ is the encoder; $f(\mathbf{I}_i), f(\mathbf{I}'_i)$ are embeddings of positive pairs from different augmentations of the same image; $\tau$ is a temperature that smooths the similarity distribution, balancing hard and easy negatives.

Unlike traditional CL methods that use only different views of the same instance as positive pairs, medical imaging CL typically constructs multiple positive pairs with similar semantics, enhancing positive diversity and producing more informative representations. Positional contrastive learning (PCL) \citep{zeng2021positional} uses anatomical similarity in volumetric images, treating spatially close slices as positive and distant ones as negative. Semantically relevant contrastive learning (SRCL) \citep{wang2022transformer} leverages a pre-trained network to extract patch features, selecting patches with the highest feature similarity as positive pairs. \cite{zeng2023additional} applies TracIn \citep{pruthi2020estimating} to estimate pairwise image similarity and identify the most similar ones as positive pairs.

CL can also be applied across multiple scales or sources to improve segmentation.  Self-supervised Anatomical eMbedding (SAM) \citet{yan2022sam} crops multiple patches with random augmentations and employs a coarse-to-fine network to extract multi-scale features. Global features are derived from the coarsest layer, while local ones are extracted from the finest layer. SAM applies CL at both global and local levels. Overlapping pixels in different patches form positive pairs, effectively capturing both global context and detailed local anatomical features. \cite{wu2021federated} adopts CL in a federated learning setting by using image features from the same client partition as positive pairs and those from different partitions across clients as negative pairs to enrich the data diversity.

Apart from using image augmentations and similarities to form positive pairs, some research explores the use of multi-modal data for this purpose. \cite{kokilepersaud2023clinically} identifies OCT images with the same clinical data values (e.g., Central Subfield Thickness (CST) \citep{prabhushankar2022olives}) as positive pairs, while differing values are used to form negative pairs. \cite{dufumier2021contrastive} enhances the diversity of positive pairs by including images of patients with similar ages as positive pairs. \cite{zhao2021unsupervised} forms positive pairs using images with similar radiomics-derived features (e.g., shape, texture) to aid in tumor classification. \citep{liang2021contrastive} trains a network to learn image features from a large number of Textual report-image pairs. These training pairs include both positive examples of images with associated reports and negative pairs that randomly pair an image with an unrelated report. Then this pre-trained model is finetuned with a small number of labeled data for downstream classification tasks. 

Unlike traditional CL requiring explicit positive and negative samples, BT-Unet \citep{punn2022bt} computes a cross-correlation matrix to measure feature similarity and redundancy between two augmented views. The diagonal elements of the matrix indicate the similarity of corresponding parts, while the off-diagonal elements represent redundancy. By maximizing the diagonal elements and minimizing the off-diagonal elements, the BT-Unet learns invariant and diverse representations. 


\textit{Clustering Method} in medical image analysis groups images or pixels based on inherent features without labels, facilitating pattern discovery. Spatial guided self-supervised clustering network (SGSCN) \citep{ahn2021spatial} enhances medical image segmentation by grouping spatially connected pixels with similar feature representations. 


\subsubsection{Joint Training UL}
Unlike pretext tasks and discriminative unsupervised learning, which rely on a single label-free task to learn representations from unlabeled data, joint training simultaneously optimizes multiple models or combines various unsupervised objectives to collaboratively enhance representation learning. \cite{haghighi2022dira} propose DiRA, a joint training framework that combines CL, image reconstruction, and adversarial learning to simultaneously learn feature representations. Adversarial learning is trained by distinguishing between the original images and the reconstructed ones. DiRA improves representation learning by preserving fine-grained details and capturing more descriptive features through the combination of three objectives. Transferable Visual Words (TransVW) \citep{haghighi2021transferable} clusters similar patient images to create a visual word \citep{sivic2003video} dataset, for example, the anatomical consistency across patients in CT/MRI scans. TransVW learns through a joint loss combining classification (predicting visual word pseudo-labels) and reconstruction (restoring perturbed visual words). \citep{zhang2021self,zhang2021sar} incorporate multi-scale information by reconstructing and predicting scale variations in images while using adversarial learning to differentiate between these scales. It combines these losses in a joint training framework.

\paragraph{Summary of Unsupervised learning} 
As shown in Table \ref{tab:unsupervised}, UL has gained popularity in recent years, largely driven by the emergence of Self-SL techniques such as SimCLR \citep{chen2020simple}, MOCO \citep{he2020momentum} (CL), and MAE \citep{he2022masked} (pretext task). While substantially reducing reliance on expert annotations, they enhance generalization and are often used as a pretraining strategy for downstream tasks (e.g., classification or segmentation). Although these methods have achieved notable success in natural image analysis, their adoption in medical imaging remains in its developmental stage. With the rise of foundation models like SAM \citep{kirillov2023segment} and ChatGPT \citep{achiam2023gpt}, extensive unlabeled-data pretraining and continual refinement are increasingly viewed as promising directions.

Currently, few other scenarios can assist UL, as it imposes the strictest requirements on the dataset, i.e., being entirely composed of unlabeled data. Nonetheless, it can still augment scenarios such as Semi-SL (Section~\ref{Semi-supervised Learning (Semi-SL)}) and inexact supervision (Section~\ref{sec: Absent Supervision}) by providing robust pre-trained representations that are subsequently fine-tuned using small labeled sets or imposing constraints.

\begin{table*}
\centering
\caption{Overview papers of Unsupervised Learning in Medical Imaging Analysis.}
\label{tab:unsupervised}
\scalebox{0.88}{
\begin{tabular}{llllll}
\hline
Reference & Category & Sub-category & body & modality & task \\ \hline
\cite{yan2023label} & Pretext Tasks & Pixel-level & Chest/Eye/Skin & X-ray/OCT/Derm & Cls \\
\cite{zhu2020rubik} & Pretext Tasks & Pixel-level & Brain & CT/MRI & Cls/Seg \\
\cite{xie2022s} & Pretext Tasks & Pixel-level & Tissue & HSI & Cls \\
\cite{jiang2022self} & Pretext Tasks & Pixel-level & Abd & CT & Seg \\
\cite{bai2019self} & Pretext Tasks & Pixel-level & Cardiac & MRI & Seg \\
\cite{holmberg2020self} & Pretext Tasks & Pixel-level & Eye & OCT & Cls \\
\cite{zhuang2019self} & Pretext Tasks & Instance-level & Brain & CT/MRI & Cls/Seg \\
\cite{ewen2021targeted} & Pretext Tasks & Instance-level & Chest & CT & Cls \\
\cite{li2021rotation} & Discriminative & CL & Eye & OCT & Cls \\
\cite{fernandez2022contrasting} & Discriminative & CL & Prostate & MRI & Cls \\
\cite{zeng2021positional} & Discriminative & CL & Cardiac & CT/MRI & Seg \\
\cite{wang2022transformer} & Discriminative & CL & Tissue & Histo & Cls \\
\cite{zeng2023additional} & Discriminative & CL & Skin/Chest & Derm/X-ray & Cls \\
\cite{yan2022sam} & Discriminative & CL & Chest & CT & Det/Seg \\
\cite{wu2021federated} & Discriminative & CL & Cardiac & MRI & Seg \\
\cite{kokilepersaud2023clinically} & Discriminative & CL & Eye & OCT & Cls \\
\cite{dufumier2021contrastive} & Discriminative & CL & Brain & MRI & Cls \\
zhao2021unsupervised & Discriminative & CL & Thyroid/Kidney & US/CT & Cls \\
\cite{punn2022bt} & Discriminative & CL & Brain/Breast/Cell & MRI/US/Histo & Seg \\
\cite{ahn2021spatial} & Discriminative & Clustering & Liver/Skin & US/Derm & Seg \\
\cite{haghighi2022dira} & Joint Training & Pixel-level + CL & Chest & X-ray & Cls/Seg \\
\cite{haghighi2021transferable} & Joint Training & Pixel-level + CL & Chest & X-ray & Cls/Seg \\
\cite{zhang2021self} & Joint Training & Pixel + Instance-level & Eye/Colon/Skin & Fundus/Endo/Derm & Seg \\
\cite{zhang2021sar} & Joint Training & Pixel + Instance-level & Brain/Pancre & MRI/CT & Seg \\ \hline
\end{tabular}}
\end{table*}


\subsection{Transductive Transfer Learning (TTL)}
\label{Transductive Transfer Learning}
Unlike UL, TTL utilizes knowledge from labeled source data to guide learning on unlabeled target data, facilitating cross-domain knowledge transfer to enhance performance \citep{pan2009survey}. TL holds significant potential in medical contexts, particularly in scenarios involving the transfer of information within the same modality or organ. This can effectively reduce the labeling costs in the target domain, especially when confronted with datasets characterized by limited or absent labels, as in the case of emerging disease outbreaks, such as Covid-19 \citep{minaee2020deep} or rare disease identification. Let $D=(\mathbb{R} ^{h \times w \times d}, P(\mathbf{I}))$ denotes the domain of the image dataset $\mathcal{D}$, and $T=(\mathcal{Y}, f(\cdot))$ represent a task that is not directly observed but can be learned from $\mathcal{D}$. TTL can be defined as: 
\begin{definition}[Transductive Transfer Learning] 
Given a labeled source dataset $\mathcal{D}_L^S$ and an unlabeled target dataset $\mathcal{D}_U^T$, transductive transfer learning aims to improve the learning of $f(\cdot)$ on $D^T$ using knowledge from $D_L^S$ and $T^S$, where $D^S \neq D^T$ and $T^S = T^T$.
\end{definition}

TTL can be summarized into instance-transfer and feature-representation-transfer, depending on the type of knowledge transferred \citep{pan2009survey}. However, instance-transfer typically requires a certain number of source-domain training instances to be related in the target domain and assumes identical conditional distributions between domains. These conditions often do not hold in real-world scenarios \citep{niu2020decade}, making instance-transfer rarely applicable in medical image analysis. Therefore, this survey primarily focuses on feature-representation TTL.

\begin{figure}[tbp]
\centering
\includegraphics[width=0.35\textwidth]{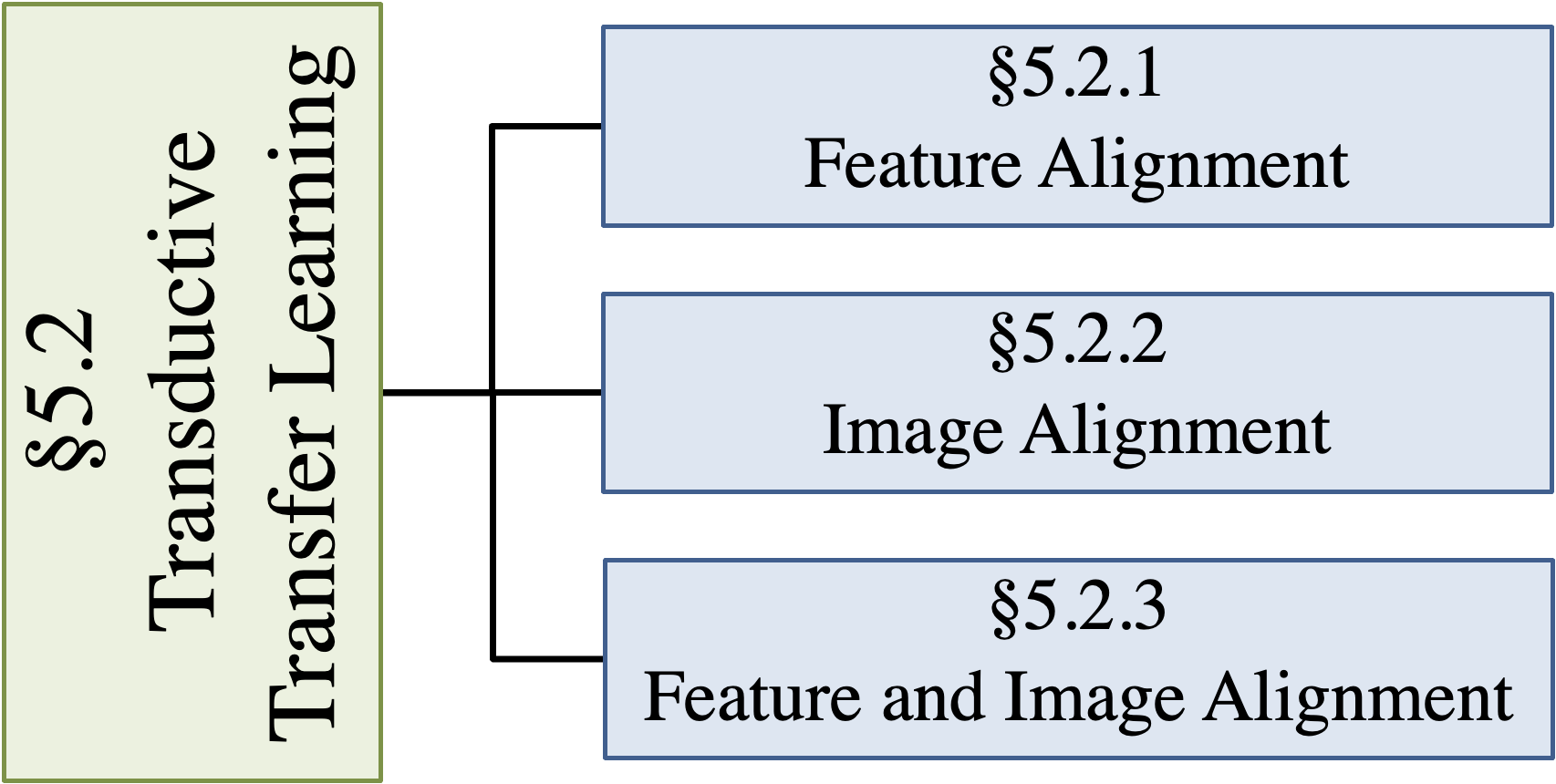}
\caption{The taxonomy of Transductive Transfer Learning.}
\label{fig:transfer_tree}
\end{figure}

Feature-representations-transfer assumes that the distributions of the source and target domains are similar \citep{niu2020decade}. The intuitive idea behind it is to minimize the distribution mismatch between the two domains by transferring or transforming feature representations into a shared space \citep{pan2009survey,niu2020decade}. Fig. \ref{fig:transfer_tree} illustrates the overall taxonomy. 

\subsubsection{Feature Alignment}
\begin{figure}[htbp]
\centering
\includegraphics[width=0.45\textwidth]{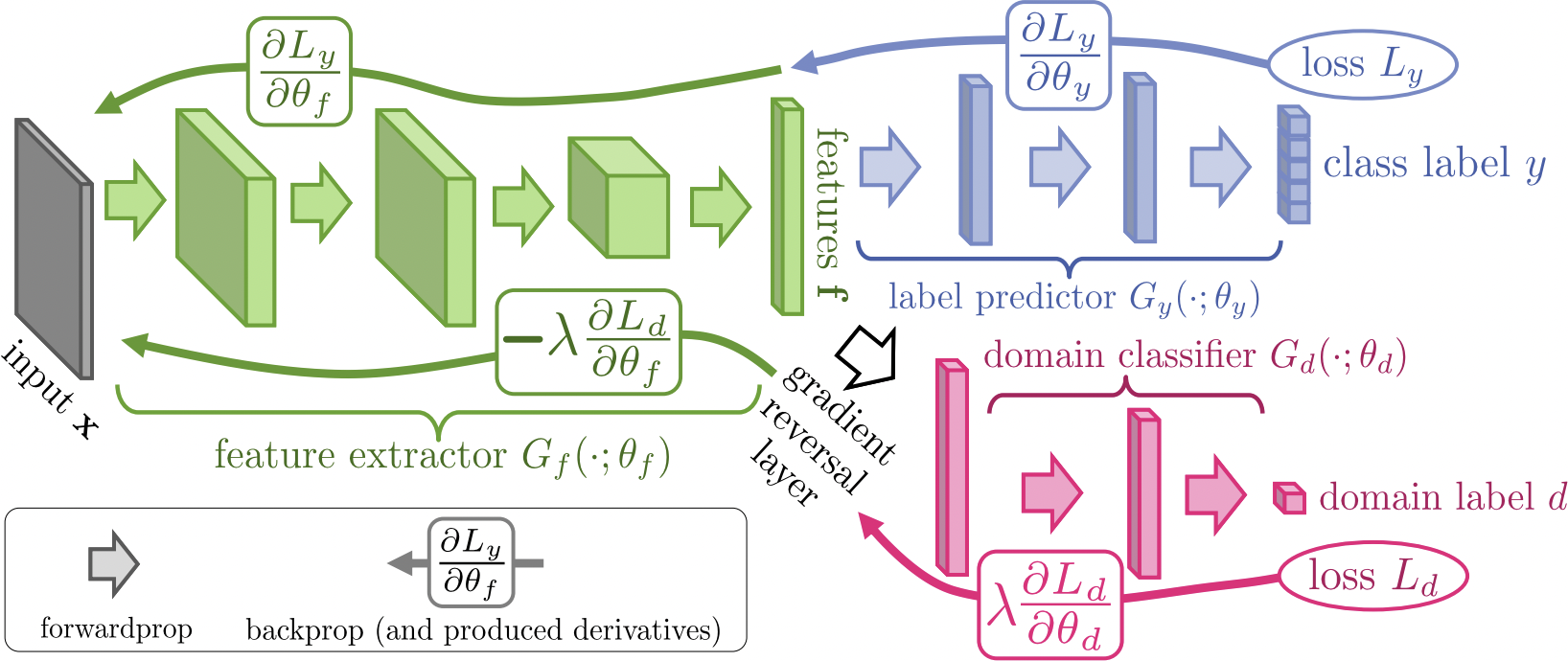}
\caption{The framework of adversarial-based domain adaptation \citep{ganin2016domain}.}
\label{fig:transfer_adv}
\end{figure}

Feature Alignment aims to learn domain-invariant features across domains. The frequently used adversarial framework \citep{ganin2016domain} consists of a feature extractor, a label predictor, and a domain classifier, with the gradient reversal layer (GRL) connecting the feature extractor and the domain classifier. During training, the label predictor optimizes the classification task on the source domain, while the domain classifier attempts to distinguish the source from the target domain. The GRL reverses the gradient from the domain classifier during backpropagation, forcing the extractor to learn domain-invariant features that minimize the discrepancy between domains. The overall framework is shown in Fig. \ref{fig:transfer_adv}.


To further leverage scarce target data in medical image segmentation, \cite{ouyang2019data} builds upon an adversarial approach by employing a VAE to further align the feature distributions of the source and target domains, forcing the feature distributions of both domains to approximate a fixed prior distribution, thereby enhancing the alignment. 

In recent years, preserving key medical features, such as organ shape and structural consistency, has emerged as a promising direction for improving domain adaptation performance in medical imaging. For example, \citep{chen2022dual,liu2023attention} enhances feature-level alignment between domains using an attention mechanism to help the model distinguish transferable regions (class-specific knowledge) from non-transferable ones (source knowledge). Additionally, in \cite{liu2023attention}, a category-weighted Maximum Mean Discrepancy (MMD) loss \citep{pan2010domain} is incorporated to minimize the impact of irrelevant source samples during training. \citep{bian2020uncertainty} attempts to merge adversarial learning with curriculum learning for anatomical segmentation, initially extracting features from the source and target domains and dynamically adjusting feature weights starting from easy cases to hard ones, according to the uncertainty map of segmentation, to boost the domain adaptation performance progressively, which then refined by domain discriminator. 

\subsubsection{Image Alignment}
Image alignment refers to the process of directly transforming images from the source domain to the target domain (or vice versa) to reduce domain shift. \citep{sanchez2022cx} selected samples from the source dataset based on similarity to the characteristics of the target dataset. Subsequently, GANs were employed to transform the selected source domain images into the target domain style to achieve image alignment. Traditional UDA methods often overlook the continuity between slices in 3D medical images. In \cite{shin2023sdc}, a 3D self-attention mechanism is employed across slices to ensure consistent style in adjacent slices, maintaining overall coherence in the volumetric data when transferring the source image to the target domain. While most methods \citep{sanchez2022cx,shin2023sdc} focus on transforming real source images to match the real target domain, \cite{mahmood2018unsupervised} addressed the reverse scenario through adversarial training, where real medical images (target) are transformed into a style similar to synthetic images (source), while retaining key medical features such as tissue structure and abnormalities. As a result, the model no longer has to deal with the complexity of real images directly but instead processes synthetic-style images, thus improving its predictive performance in real-world scenarios.


\subsubsection{Feature and Image Alignment}
Approaches that attempt to align domains at the image or feature level fail to consider the unified relationships between cross-modal images and their corresponding features. Therefore, researchers started to discover the combination of feature and image alignment in the same framework. For example, \cite{li2021attent} employs the CycleGAN combined with an attention mechanism to enhance the semantic and geometric consistency of the target organs during the image style translation process. Furthermore, consistency is constrained in image segmentation between labeled images in the source domain and unlabeled images in the target domain for multi-organ segmentation, to maintain the key medical features, which enhances feature alignment. Similarly, \citep{su2023mind} utilizes style transfer to globally align source domain images with the target domain in style, yet preserving their inherent content. To achieve this, feature and label domain discriminators are used to correct subtle local discrepancies between the source and target domains. In \cite{wang2022cycmis,wang2023edrl}, image representations are decomposed into two parts: a domain-invariant content and a domain-specific appearance vector. By combining content vectors derived from the source domain with appearance vectors randomly sampled from the target domain, a transformation of appearances is achieved. Adversarial learning is used for feature-level alignment as usual.

\begin{table*}
\centering
\caption{Overview papers of Transductive Transfer Learning in Medical Imaging Analysis.}
\label{tab:transfer}
\scalebox{0.88}{
\begin{tabular}{lllllll}
\hline
Reference & Category & Source body & Source modality & Target body & Target modality & Task \\ \hline
\cite{xie2022unsupervised} & Feature Align & Abd/Brain/Cardiac & MRI/CT & Abd/Brain/Cardiac & MRI/CT & Seg \\
\cite{chen2022dual} & Feature Align & Brain/Cardiac & CT/MRI & Brain/Cardiac & MRI/CT & Seg \\
\cite{ouyang2019data} & Feature Align & Cardiac & MRI & Cardiac & CT & Seg \\
\cite{bian2020uncertainty} & Feature Align & Cardiac/Eye & MRI/OCT & Cardiac/Eye & CT/OCT & Seg \\
\cite{liu2023attention} & Feature Align & Chest & X-ray & Chest & X-ray & Cls \\
\cite{xian2023unsupervised} & Feature Align & Abd & CT/MRI & Abd & CT/MRI & Seg \\
\cite{shin2023sdc} & Image Align & Brain/Cardiac & MRI & Brain/Cardiac & MRI/CT & Seg \\
\cite{mahmood2018unsupervised} & Image Align & Colon & Endo & Colon & Endo & Reg \\
\cite{sanchez2022cx} & Image Align & Chest & X-ray & Chest & X-ray & Cls \\
\cite{li2021attent} & Feature + Image Align & Abd & MRI/CT & Abd & MRI/CT & Seg \\
\cite{su2023mind} & Feature + Image Align & Abd/Cardiac & CT/MRI & Abd/Cardiac & CT/MRI & Seg \\
\cite{wang2023edrl} & Feature + Image Align & Cardiac & CT/MRI & Cardiac & MRI & Seg \\
\cite{wang2022cycmis} & Feature + Image Align & Cardiac & MRI & Cardiac & MRI & Seg \\
\cite{xia2020uncertainty} & Feature Align (Semi-SL) & Abd/Pancre & CT & Abd/Pancre & CT & Seg \\
\cite{zhao2021mt} & Feature Align (Semi-SL) & Cardiac & MRI & Cardiac & CT & Seg \\
\cite{liu2022margin} & Feature Align (UL) & Cardiac & MRI/CT & Cardiac & MRI/CT & Seg \\
\cite{sharma2022mani} & Feature Align (UL) & Tissue & Histo & Cell & Histo & Seg \\
\cite{kuang2023mscda} & Feature Align (UL) & Breast & MRI & Breast & MRI & Seg \\
\cite{wang2022embracing} & Feature Align (UL) & Brain & MRI & Brain & MRI & Cls \\ \hline
\end{tabular}}
\end{table*}

\subsubsection{Integration with other techniques}
TTL primarily focuses on leveraging knowledge from the source domain and transferring it to assist learning in the target domain. However, due to the significant differences between domains, TTL often leads to suboptimal results, limiting the model's generalization in the target domain, and leading to the trend toward integrating it with other learning paradigms.


\textit{Semi-SL} methods like self-training and consistency regularization can support domain adaptation and target-domain learning by generating pseudo labels in unlabeled target datasets. Uncertainty-Aware Multi-View Co-Training (UMCT) \citep{xia2020uncertainty} trains on different views of the same data to capture richer representations. By maintaining consistent predictions across domains, it reduces the source and target gap. Additionally, pseudo labels assist in target-domain learning, making such methods effective in both UDA and Semi-SL settings. \cite{zhao2021mt} introduces a dual-teacher model to extract semantic knowledge and anatomical structure from the source and target domains, respectively. Each student model learns from its domain-specific teacher, while cross-domain consistency is enforced between student predictions to promote structural alignment (e.g., consistent heart shape across modalities). Similarly, \cite{xie2022unsupervised} employs a shared anatomical encoder and two domain-specific encoders to disentangle anatomical and modality features, incorporating shape constraints. The generated pseudo labels are used to adaptively enhance segmentation performance in the target domain.

\textit{Self-SL} can further facilitate feature learning and transfer by incorporating prior medical knowledge and structural constraints, as discussed in earlier sections. This direction has shown growing promise in recent studies. \citep{xian2023unsupervised} involves CL, where slices from the same patient are regarded as positive pairs to guide feature learning. Segmentation from different domains is constrained by ensuring the domain discriminator cannot distinguish their sources. Similarly, to solve nuclei cross-domain segmentation, \cite{sharma2022mani} computes the similarity between nuclei features from the source and target domains. Features with high similarity are treated as positive pairs, helping maximize mutual information across domains. In \cite{liu2022margin}, labeled source data is used to generate class prototypes representing feature centers. Target unlabeled data are pseudo-labeled, and samples from the same class (regardless of domain) form positive pairs; otherwise, they are negatives. This aligns target features with source prototypes by maximizing similarity. \cite{kuang2023mscda} enhances TTL not only through global domain alignment but also via detailed semantic guidance. It employs CL integrating pixel-to-pixel (aligning pixel-level features between source and target), pixel-to-centroid (comparing pixels with class centroids), and centroid-to-centroid (aligning centroids between domains) to reduce semantic gap. Additionally, \citep{wang2022embracing} uses source data for pretraining with auxiliary demographic prediction (e.g., age, gender), aiding generalization by capturing cross-domain commonalities.

\paragraph{Summary of Transductive Transfer Learning}
As shown in Table~\ref{tab:transfer}, TTL has been studied across diverse source-target configurations, including identical or distinct body parts and imaging modalities. Many studies focus on domain adaptation via feature alignment, particularly using adversarial methods to handle unlabeled target datasets. By aligning feature distributions between source and target domains, models trained on a labeled source domain can effectively generalize to a target domain lacking annotations. 

Nonetheless, TTL still faces key challenges, including substantial domain gaps, such as variations in scanner protocols or inherent differences in patient populations across geographic regions. Additionally, leveraging structural priors (e.g., the relatively consistent morphology of the specific organs) has emerged as a promising avenue to mitigate cross-domain inconsistencies. Recent studies further demonstrate a trend of integrating semi- or unsupervised learning components to refine feature representations.


\section{Discussion}
\label{sec: Discussion}
In this section, we present further analysis and discussion to complement the prior taxonomy. Section~\ref{Scenario Trends} outlines the development trends of various research scenarios, including AL, Semi-SL, ISL, UL, and TTL. Then the most common tasks, target organs, imaging modalities, and datasets across these paradigms are analyzed in Section~\ref{Task and Data Overview Across Learning Scenarios}. Finally, Section~\ref{Future direction} synthesizes the above findings to highlight practical challenges, relationships between learning scenarios, and emerging research directions.

\subsection{Scenario Trends}
\label{Scenario Trends}
\begin{figure}[tbp]
\centering
\includegraphics[width=0.48\textwidth]{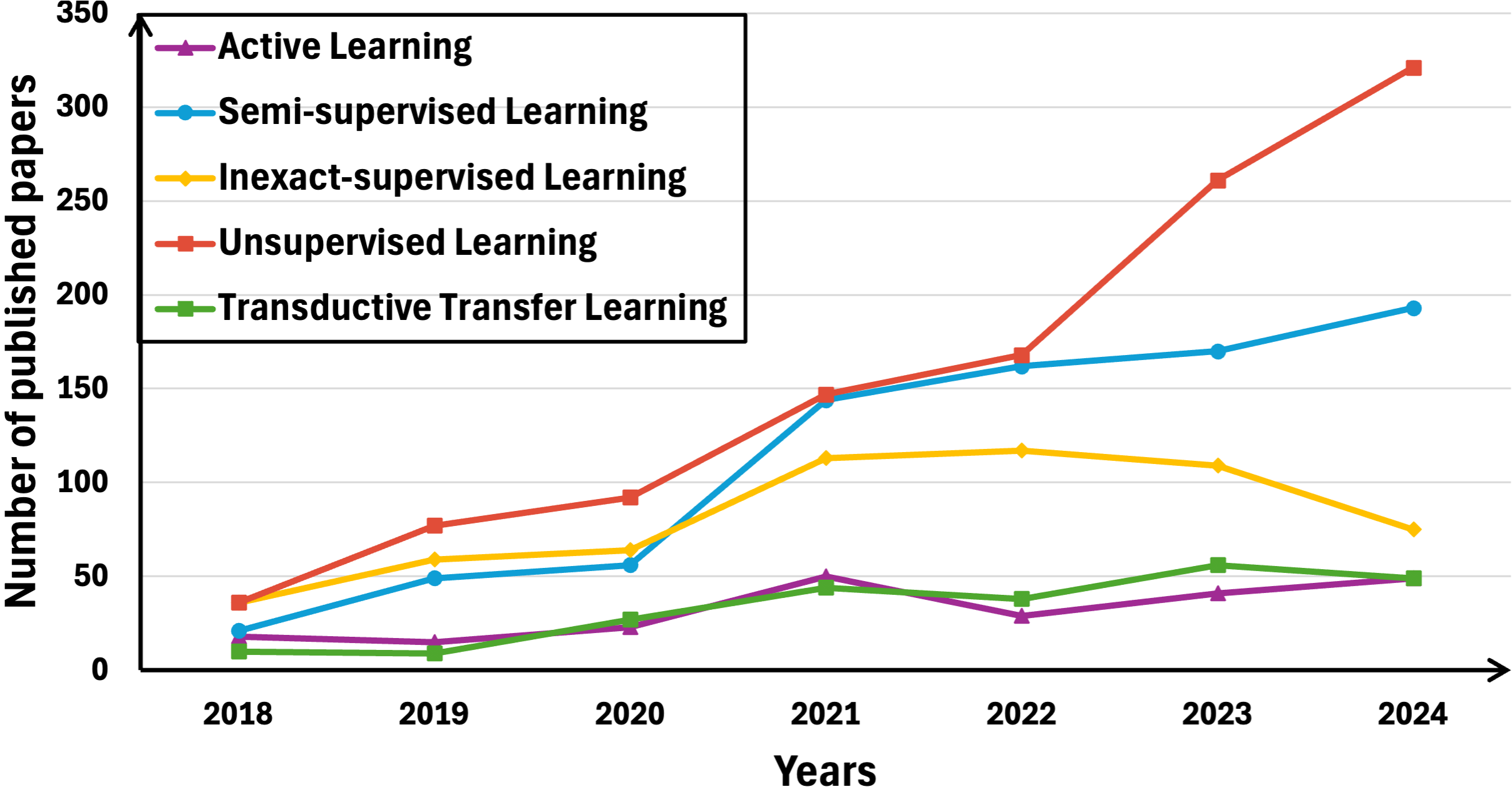}
\caption{An overview trend of all techniques considered in this paper.}
\label{fig:trend}
\end{figure}

We begin by analyzing the temporal trends across learning scenarios, including AL, Semi-SL, ISL, UL, and TTL, by tracking the number of published papers per year in each category. 

As shown in Fig.~\ref{fig:trend}, from 2018 to 2024, UL has consistently held the most prominent position, particularly due to the rise of self-supervised learning (Self-SL). A significant increase in the number of related publications has been observed since 2022, a trend likely to continue with the growing adoption of large foundation models \citep{kirillov2023segment,achiam2023gpt} and pre-training strategies.

Semi-SL and ISL have emerged as two prominent areas of research. Semi-SL, in particular, has shown steady growth, with a sharp surge in 2021, reaching nearly 150 publications. Its diverse approaches suggest sustained interest in the field. Conversely, inexact supervised learning has seen a slight decline, likely due to a lack of significant breakthroughs beyond traditional methods like Multiple-Instance Learning (MIL), attention mechanisms, and pseudo-labeling techniques.


In contrast, TTL and AL-related publications have been underwhelming in recent years. Since TL mainly addresses domain adaptation, only a small number of papers deal with scenarios where the target dataset has no labeled samples (TTL), as a way to reduce labeling cost. Thus, a low number of TL-related publications in this context is expected. Perhaps due to AL's inconsistency in stability across different methods and tasks, and its reliance on human experts for annotation, has struggled to establish itself in practical applications. As a result, these two fields are not expected to experience substantial expansion in the short term.

\begin{table}[]
\centering
\caption{Top-3 bodies, modalities, tasks, and corresponding case counts in each learning scenario. The total number of reviewed papers per scenario is as follows: AL (28), Semi-SL (126), ISL (32), UL (50), and TTL (70).}
\label{tab:scenario}
\scalebox{0.88}{
\begin{tabular}{llll}
\hline
Scenerio & Body \textbar{} Num & Modality \textbar{} Num & Task \textbar{} Num \\ \hline
AL & Chest \textbar{} 8 & X-ray \textbar{} 8 &  Seg \textbar{} 17 \\
 & Brain \textbar{} 5 & CT \textbar{} 8 & Cls \textbar{} 14\\
 & Skin \textbar{} 3 & MRI \textbar{} 7 & Det \textbar{} 1 \\ \hline
Semi-SL & Cardiac \textbar{} 44 & MRI \textbar{} 60 & Seg \textbar{} 88 \\
 & Chest \textbar{} 21 & CT \textbar{} 50 & Cls \textbar{} 34 \\
 & Brain \textbar{} 19 & X-ray \textbar{} 16 & Det \textbar{} 5 \\ \hline
ISL & Cell \textbar{} 6 & Histo \textbar{} 11 & Seg \textbar{} 14 \\
 & Tissue \textbar{} 5 & MRI \textbar{} 11 & Cls \textbar{} 10 \\
 & Chest/Cardic \textbar{} 5 & CT \textbar{} 5 & Det \textbar{} 8 \\ \hline
UL & Brain \textbar{} 18 & MRI \textbar{} 28 & Seg \textbar{} 39 \\
 & Chest \textbar{} 16 & CT \textbar{} 27 & Cls \textbar{} 29 \\
 & Eye \textbar{} 12 & OCT \textbar{} 12 & Det \textbar{} 16 \\ \hline
TTL & Cardiac \textbar{} 28 & MRI \textbar{} 33 & Seg \textbar{} 38 \\
 & Abd \textbar{} 11 & CT \textbar{} 17 & Cls \textbar{} 10 \\
 & Brain \textbar{} 11 & Fundus \textbar{} 3 & Det \textbar{} 2 \\ \hline
\end{tabular}}
\end{table}

\begin{table*}[h]
\centering
\caption{Summary of representative medical imaging datasets used in various learning scenarios. Scale indicates typical usage in literature (* denotes variation by year). }
\label{tab:datasets}
\scalebox{0.88}{
\begin{tabular}{llllllll}
\hline
Dataset & Scenarios & Body & Modality & Dim & Task & Label Type & Scale \\ \hline
BraTS \citep{bakas2018identifying}& AL/Semi-SL/UL/TTL & Brain & MRI & 3D & Seg & Pixel-level tumor & 285*\\
ChestX-ray14 \citep{wang2017chestx}& AL/Semi-SL/ISL/UL & Chest & X-ray & 2D & Cls & 14-class image-level & 112k \\
ISIC \citep{codella2019skin}& AL & Skin & Derm & 2D & Cls & 7-class image-level & 13k* \\
ACDC \citep{bernard2018deep}& Semi-SL/TTL & Cardiac & MRI & 3D & Seg & Pixel-level structure & 100 \\
CAMELYON16 \citep{bejnordi2017diagnostic} & ISL & Breast & Histo & 2D & Det/Seg & Image-level/sparse ROI & 400 \\
GlaS \citep{sirinukunwattana2017gland} & ISL & Colon & Histo & 2D & Seg & Coarse pixel-level gland & 165 \\
OCT2017 \citep{kermany2018identifying} & UL & Retina & OCT & 2D & Cls & 4-class image-level & 84k \\
CHAOS \citep{kavur2021chaos}& TTL & Abd & CT/MRI & 2D/3D & Seg & Pixel-level multi-organ & 40 \\ \hline
\end{tabular}}
\end{table*}

\subsection{Task and Data Overview Across Learning Scenarios}
\label{Task and Data Overview Across Learning Scenarios}
To illustrate how each learning scenario aligns with specific data characteristics and tasks, we integrate insights from Table \ref{tab:scenario} (showing top body parts, imaging modalities, and tasks in each scenario) and Table \ref{tab:datasets} (highlighting representative public datasets). It is worth noting that these statistics are derived from a curated list of papers selected based on target conferences and journals, as detailed in the supplementary material.

\subsubsection{Overall Observations}
From Table \ref{tab:scenario}, across nearly all scenarios, segmentation tasks dominate (e.g., Semi-SL with 88 segmentation papers, TTL with 38). This reflects the high annotation cost in medical images, especially 3D (e.g., MRI, CT), where a single scan can contain hundreds of slices requiring manual labeling. Correspondingly, MRI and CT consistently appear as top imaging modalities (Semi-SL: MRI 60/CT 50; TTL: MRI 33/CT 17), which can be also attributed to the availability of large-scale public benchmarks. In terms of anatomical coverage, these modalities are the primary choices for imaging the brain, cardiac, and abdomen, which aligns with the dominance of datasets like BraTS and ACDC, as listed in Table~\ref{tab:datasets}, which serve as classic benchmarks for 3D segmentation in label-efficient learning.

\subsubsection{Scenario-Specific Insights}
\textit{AL} shows a balanced distribution between segmentation (17 papers) and classification (14). For classification, the low annotation cost and operational simplicity make iterative sample querying both feasible and effective.
For segmentation, pixel-level labeling remains much more costly and labor-intensive, especially in 3D images. However, unlike other scenarios such as Semi-SL or TTL, where segmentation tasks significantly outnumber classification, AL does not exhibit such a strong preference. This is primarily due to the pixel-level granularity of segmentation labels. Identifying the most informative regions for annotation in dense 2D/3D images remains a fundamental challenge. Moreover, random selection often serves as a surprisingly strong baseline.
Nonetheless, recent studies \citep{ma2024breaking,li2023hal} have actively addressed this issue, proposing more robust and adaptive query methods tailored to dense prediction tasks. Frequently used classification datasets include ChestX-ray14 (over 112k images) and ISIC-2018 (13k dermoscopic images), both of which offer large unlabeled pools ideal for AL sampling. In terms of performance, AL methods can often achieve results comparable to fully supervised models using only 20–50\% of the labeled data \citep{smailagic2018medal,li2023hal,ma2024breaking}, depending on dataset characteristics and task difficulty.

\textit{Semi-SL} is predominantly applied to segmentation tasks (88 out of 126 papers), especially in the 3D domain, where voxel-wise annotations are notoriously expensive and time-consuming. Semi-SL is particularly well-suited for such scenarios as it allows models to leverage abundant unlabeled slices or volumes, significantly reducing the reliance on dense manual labels. Prominent datasets such as BraTS and ACDC are frequently used benchmarks in this setting. For example, BraTS 2018 provides 285 3D MRI volumes, corresponding to roughly 45,000 slices, and is commonly employed across a variety of paradigms including AL, Semi-SL, and UL. Under the same limited annotation budget, many Semi-SL approaches report 2–5\% improvements in Dice or IoU scores over fully supervised baselines \citep{wu2022minimizing,wang2022rethinking,sedai2019uncertainty}. However, this performance gap tends to narrow as the size of the labeled dataset increases, illustrating that Semi-SL is particularly valuable when labeled data is limited.

\textit{ISL} exhibits a balanced task distribution across segmentation, ROI classification, and detection tasks, compared to other learning paradigms. Notably, ISL is particularly well-suited for histopathology applications, where whole slide images (WSIs) are extremely large and accurate pixel-level annotation is highly time-consuming. Representative datasets like CAMELYON16 and GlaS (in Table \ref{tab:datasets}) are widely adopted in this context, owing to their large image size and partially annotated structures, such as image-level or sparse ROI annotations, are available. Although ISL-based methods may not always match the performance of fully supervised approaches, they often achieve within 3–10\% of the best Dice or detection scores \citep{wang2020weakly,li2022deep}, while drastically reducing annotation effort.

In \textit{UL} settings, segmentation (39 papers) and classification (29) are the most common tasks. Large-scale datasets are often chosen as pretraining sources due to their abundance of unlabeled images. For example, Brain MRI datasets such as BraTS 2018 provide 285 3D volumes (45k slices), ChestX-ray14 contains 112k images, and OCT2017 offers 84k retinal images. These datasets are ideal for Self-SL, which aims to extract generalizable representations from unlabeled data. Empirically, 2–10\% performance gains (e.g., in Dice for segmentation or AUC for classification) are commonly reported compared to random initialization \citep{zhu2020rubik,jiang2022self,zhang2021sar}. Notably, larger datasets tend to yield greater benefits from unsupervised pretraining.

For \textit{TTL}, segmentation remains the dominant task, which primarily aims to bridge domain gaps across modalities or sources. This makes TTL especially well-suited for multi-modality or cross-domain datasets. A representative example is the CHAOS dataset, which contains both MRI and CT scans of the same abdominal organs. This makes it an ideal benchmark for validating domain adaptation and cross-modality alignment techniques. In practice, unsupervised domain adaptation on unlabeled target datasets can yield 20–70\% performance improvements compared to models without source-domain adaptation \cite{chen2022dual, wang2023edrl}. Notably, this is achieved without requiring any labeled data in the target domain, making it a powerful solution for reducing annotation costs. However, the effectiveness strongly depends on the similarity between the source and target domains, including factors such as imaging protocols and patient demographics.

In summary, each learning scenario offers unique advantages tailored to specific data characteristics and task requirements. When selecting an appropriate strategy, one should first consider the degree of label availability. For instance, datasets with partial precise labels and a large number of unlabeled data are best suited for incomplete supervision paradigms, such as AL or Semi-SL. In contrast, datasets with coarse annotations (e.g., image-level or scribble-level labels) are better addressed by ISL. When dealing with multi-modal or cross-organ settings, TTL provides an effective means to reduce annotation costs by transferring knowledge across domains. Specifically, within incomplete supervision, the choice between AL and Semi-SL often depends on the task: AL is more commonly applied to classification problems due to its ease of label acquisition, while Semi-SL has shown greater promise in segmentation tasks, especially in 3D medical imaging.

Furthermore, combining multiple scenarios (e.g., AL and Semi-SL) has emerged as a promising strategy, leveraging their complementary strengths. A more detailed discussion on scenario integration and hybrid paradigms can be found in Section \ref{Relation between learning scenarios and More advanced hybrid method}.

\subsection{Future direction}
\label{Future direction}

This section highlights several significant developments in the field of medical image analysis. Based on the characteristics of medical imaging, we have summarized the challenges faced by the current varying degrees of label learning in medical image analysis in Section \ref{Challenge of not-so-supervised learning scenario}. These are the key issues that require attention. Section \ref{Relation between learning scenarios and More advanced hybrid method} analyzes the relationships between these methods and their strengths and weaknesses, to facilitate better integration. In Section \ref{Similar and Key idea behind scenario}, we will summarize the current high-quality ideas that are common across different scenarios. By doing so, we hope to encourage the generation of new ideas and facilitate further development in the field.

\subsubsection{Challenges of scenario with varying label availability}
\label{Challenge of not-so-supervised learning scenario}
Influenced by the unique characteristics of medical images, the scenario under varying label availability presents several challenges. In this section, we summarize these challenges to assist potential researchers in better understanding the current issues in the field and provide potential solutions.

\begin{enumerate}
  \item \textit{Imbalanced Data} \citep{zhu2018class}: Models trained on imbalanced data often favor frequent categories, which compromises generalization, especially in scenarios with limited or no labeled data. Several approaches have been proposed to mitigate this issue, such as oversampling underrepresented classes using AL, augmenting rare-class samples via generative models, or employing class-balanced loss functions like focal loss \citep{lin2017focal}.

  \item \textit{Multi-Label Learning} \citep{choi2017generating}: Medical images often contain multiple co-occurring diseases, making multi-label classification natural yet challenging. Ambiguous boundaries and semantic label overlap hinder prediction, especially under limited supervision. Recent solutions include multi-dimensional feature fusion and confidence-based classification strategies, which offer more flexible and accurate representations of disease presence in multi-label settings.

  \item \textit{Domain Variance} \citep{li2020domain}: A dataset may have samples from different hospitals or machines, resulting in domain variance. This dissimilarity can lead to inconsistencies in labels or label quality in Semi-SL or UL settings. TL techniques are specifically designed to address this problem by adapting models to different domains and mitigating domain shift issues.


  \item \textit{Interpretability issue} \citep{ahmad2018interpretable}: In the medical domain, interpretability is critical for building trust and enabling informed clinical decisions. This becomes even more challenging under limited or weak supervision, where model behavior is harder to validate. Common methods include sample distribution visualization (e.g., in AL, to display queried data) and heatmaps that highlight salient regions in images, helping experts understand the model’s attention. Additionally, providing uncertainty estimates enables clinicians to assess the model’s confidence and make better-informed decisions.



  \item \textit{Performance limitation}: Under varying degrees of label availability, the model performance is often compromised due to noise, and uncertainty. Inaccurate-supervised learning \citep{zhou2018brief} can help identify and filter out potential noisy labels, ensuring better data reliability. Additionally, noise-robust loss functions and specialized models can mitigate the impact of label noise. Also, combining multiple methods from different scenarios can further help solve this issue, as illustrated in Section \ref{Relation between learning scenarios and More advanced hybrid method}. 
\end{enumerate}

\subsubsection{Interconnections Among Learning Paradigms}
\label{Relation between learning scenarios and More advanced hybrid method}
The scenarios presented in the previous sections all aim to decrease labeling costs. In this section, we analyze the similarities, differences, and relationships between these methods to inspire better combinations.


\textit{AL} involves selecting informative samples for labeling, with the aid of human participation. For the remaining large amount of unlabeled data, the scenario can be considered as UL or Semi-SL, achieving a mutually beneficial effect. For example, by leveraging unlabeled data to pre-train the model, Self-SL enhances the model's feature extraction abilities, allowing it to perform better even with minimal labeled data. In addition, Self-SL can effectively address the cold-start problem in AL, where obtaining an initial labeled dataset is challenging due to the suboptimal performance of the untrained model. 

\textit{Semi-SL} shares similarities with AL in that it uses limited labeled data and a large number of unlabeled data without human involvement. Pseudo-label generation is one of the most popular methods, but can introduce noise into the learning process. Hence, the incorporation of external constraints, such as AL involving interactive learning, where Semi-SL generates pseudo labels and experts participate in refining and correcting them, or Self-SL employing generalized data learning, can provide additional assistance to the process. 

\textit{ISL} is distinct from the previous two types of learning in that it utilizes a coarse label as guidance for all samples but lacks a label suitable for intense target tasks like segmentation. One learning approach is to explore data information directly. The full label predicted on this basis can be viewed as a pseudo-label, which shares similarities with Semi-SL. Combining the ideas of these two methods could result in improved predictions, such as using all coarse labels and a small proportion of full labels for a sample to achieve better performance. This could still reduce labeling costs to a large extent. Similarly, AL can introduce interactive learning to continuously optimize predictions for better results.

\textit{UL} also learns from unlabeled data but has broader applicability than previous methods, as it requires no labeled data. This flexibility allows it to be applied across various learning scenarios. We anticipate further advancements in Self-SL techniques and their integration into other paradigms.

\textit{TTL} leverages knowledge from the source domain to aid target dataset learning. Simple fine-tuning on pre-trained models in TL can be used in other scenarios. For example, multiple rounds of model training fine-tuning to reduce training time consumption in AL \citep{Zhou2017FineTuningCN}. Semi-SL and Self-SL can enhance domain-level semantic consistency by generating pseudo labels or ensuring that the shape of the heart remains consistent across different modalities. 

\subsubsection{Similar and Key idea behind scenario}
\label{Similar and Key idea behind scenario}
Although scenarios have their unique approaches, we can identify several key ideas that they share, which can be observed across various methods and possess good generality. 

\begin{enumerate}
  \item \textit{Data augmentation} creates new training samples by applying transformations like flips or crops to existing data. This technique helps the model learn feature invariance, making it more robust to minor input changes. Additionally, it reduces the risk of overfitting by increasing the diversity of the limited training set, allowing the model to generalize better to unseen data.

 \item \textit{Consistency regularization} widely adopted strategy across various Semi-SL frameworks, including co-training, self-ensembling, and Mean Teacher (MT) methods. These approaches typically apply data augmentation to generate perturbed versions of the same input, and enforce prediction consistency across them. This encourages the model to produce stable outputs under input perturbations, thereby enhancing its robustness and generalization when labeled data is limited.

 \item \textit{Contrastive learning}, similar to consistency regularization, aims to capture the general data distribution in the latent space by leveraging the relationship between similar and dissimilar samples. In this approach, similar samples, known as positive pairs, are encouraged to have closer representations, while different samples, referred to as negative pairs, are pushed apart.

 \item \textit{Ensemble learning} can be achieved in various ways, including feature-level ensemble, which combines features from multiple layers; data-level ensemble, which leverages data augmentation techniques to enhance diversity; and network or task-level ensemble, which involves utilizing ensemble models or multi-task learning approaches to learn a more robust feature representation. In Semi-SL settings, ensemble techniques are often combined with consistency regularization to further stabilize training and enhance generalization.

 \item \textit{Adversarial-Based Training}: This approach aims to improve model robustness and generalization by introducing an adversarial objective during training. A typical example is GANs, where a generator and discriminator are trained in a minimax game to synthesize realistic samples or align feature distributions. Variants of adversarial training have been applied across multiple learning scenarios, including AL, Semi-SL, and TTL, particularly for domain adaptation and data augmentation purposes.

 \item \textit{Attention mechanism} is commonly used in the model construction, from the first appearance of attention blocks in CNNs to the development of ViT. These methods use attention to learn important features in various domains, as a way of interpretability. This can be also used in ISL to indicate the potential target area. 
\end{enumerate}


\subsubsection{Potential Research Direction}
This section highlights several significant advancements and ongoing trends in medical image analysis under varying degrees of labels.

Firstly, it is crucial to note the unique characteristics of medical images when applying analytical techniques such as imbalanced class distributions, multi-label problems, and domain variability, which are pivotal issues that require targeted solutions. Overcoming these limitations is essential for developing more robust, generalizable models, particularly when the fully exact label is not accessible, as discussed in Section \ref{Challenge of not-so-supervised learning scenario}. 

Secondly, as mentioned in Section \ref{Relation between learning scenarios and More advanced hybrid method}, there is a growing interest in developing integrated methods that involve multiple learning scenarios and methods. Future research should focus on designing comprehensive frameworks that combine techniques like self-SL, semi-SL, and AL. Such hybrid approaches can leverage the strengths of each method, allowing for significant reductions in labeled data requirements. 

Moreover, the generation of key ideas is essential for significant breakthroughs in all areas, which is not limited to the medical domain. We hope to see the generation of novel ideas to facilitate further development in the field like contrastive learning demonstrated in Section \ref{Similar and Key idea behind scenario}. 

Additionally, the rise of large-scale pre-trained models like SAM \citep{kirillov2023segment} and ChatGPT \citep{achiam2023gpt} offers new ways to reduce labeling costs by leveraging their ability to learn rich feature representations from a large number of data. Fine-tuning these models with minimal labeled data can achieve competitive performance, and AL can optimize sample selection for fine-tuning. Future work should explore how to best transfer knowledge to domain-specific tasks with limited data.

Multi-modality has become a key topic in recent years, offering a powerful approach to representation learning by integrating information from diverse data sources. In medical imaging, it naturally occurs across modalities such as CT, MRI, and PET. Beyond images, metadata like patient demographics (e.g., age, gender) and physician reports can further improve learning. For example, integrating this metadata with natural language processing (NLP) techniques can facilitate the development of comprehensive multi-modal models that learn from both visual and textual information like Contrastive Language–Image Pre-training (CLIP) \citep{radford2021learning}. Additionally, cross-modal consistency regularization, such as CL applied to paired data from different modalities of the same patient, promotes robust feature alignment and improves generalization. These strategies not only enhance the model's ability to capture complex inter-modal relationships but also better exploit unlabeled data by incorporating richer information. This holds promise for significantly reducing the need for costly manual annotations while achieving strong performance in medical image analysis.

\section{Conclusion}
\label{sec: Conclusion}
In this survey, we comprehensively reviewed approximately 600 studies addressing deep learning techniques for medical imaging under varying degrees of supervision, including incomplete (active and semi-supervised learning), inexact, and absent (unsupervised and transductive transfer learning) scenarios. We provide an in-depth overview of influential techniques from key papers, highlighting methodological developments within each scenario. By covering a broad range of scenarios and clearly outlining the distinctions, commonalities, and practical impacts of these approaches, this survey offers researchers structured guidance for selecting appropriate methods tailored to specific medical imaging applications. Furthermore, we identify open challenges and outline critical future directions, with particular emphasis on the integration of multi-modal data and the effective utilization of large-scale foundation models.



\setlength{\bibsep}{0pt}  
\renewcommand{\bibfont}{\small}
\bibliographystyle{model2-names.bst} \biboptions{authoryear}


\end{document}